# The NLP Cookbook: Modern Recipes for Transformer based Deep Learning Architectures

**SUSHANT SINGH[1], AND AUSIF MAHMOOD[2]**
Department of Computer Science & Engineering, University of Bridgeport, Connecticut, CT 06604, USA

Corresponding author: Sushant Singh (sushants@my.bridgeport.edu)

**ABSTRACT** In recent years, Natural Language Processing (NLP) models have achieved phenomenal success in linguistic and semantic tasks like text classification, machine translation, cognitive dialogue systems, information retrieval via Natural Language Understanding (NLU), and Natural Language Generation (NLG). This feat is primarily attributed due to the seminal Transformer architecture, leading to designs such as BERT, GPT (I, II, III), etc. Although these large-size models have achieved unprecedented performances, they come at high computational costs. Consequently, some of the recent NLP architectures have utilized concepts of transfer learning, pruning, quantization, and knowledge distillation to achieve moderate model sizes while keeping nearly similar performances as achieved by their predecessors. Additionally, to mitigate the data size challenge raised by language models from a knowledge extraction perspective, Knowledge Retrievers have been built to extricate explicit data documents from a large corpus of databases with greater efficiency and accuracy. Recent research has also focused on superior inference by providing efficient attention to longer input sequences. In this paper, we summarize and examine the current state-of-the-art (SOTA) NLP models that have been employed for numerous NLP tasks for optimal performance and efficiency. We provide a detailed understanding and functioning of the different architectures, a taxonomy of NLP designs, comparative evaluations, and future directions in NLP.

## I. INTRODUCTION

Natural Language Processing (NLP) is a field of Machine Learning dealing with linguistics that builds and develops Language Models. Language Modeling (LM) determines the likelihood of word sequences occurring in a sentence via probabilistic and statistical techniques. Since human languages involve sequences of words, the initial language models were based on Recurrent Neural Networks (RNNs). Because RNNs can lead to vanishing and exploding gradients for long sequences, improved recurrent networks like LSTMs and GRUs were utilized for improved performance. Despite enhancements, LSTMs were found to lack comprehension when relatively longer sequences were involved. This is due to the reason that the entire history known as a context, is being handled by a single state vector. However, greater compute resources lead to an influx of novel architectures causing a meteoric rise of Deep Learning [1] based NLP models.

The breakthrough Transformer [2] architecture in 2017 overcame LSTM's context limitation via the Attention mechanism. Additionally, it provided greater throughput as inputs are processed in parallel with no sequential dependency. Subsequent launches of improved Transformer based models like GPT-I [3] and BERT [4] in 2018 turned out to be a climacteric year for the NLP world. These architectures were trained on large datasets to create pre-trained models. Thereafter transfer learning was used to fine-tune these models for task-specific features resulting in significant performance enhancement on several NLP tasks [5],[6],[7],[8],[9],[10]. These tasks include but are not limited to language modeling, sentiment analysis, question answering, and natural language inference.

This accomplishment lacked the transfer learning's primary objective of achieving high model accuracy with minimal fine-tuning samples. Also, model performance needs to be generalized across several datasets and not be task or dataset-specific [11],[12],[13]. However, the goal of high generalization and transfer learning was being compromised as an increasing amount of data was being used for both pre-training and fine-tuning purposes. This clouded the decision whether greater training data or an improved architecture should be incorporated to build a better SOTA language model. For instance, the subsequent XLNet [14] architecture possessed novel yet intricate language modeling, that provided a marginal improvement over a simplistic BERT architecture that was trained on a mere ~10% of XLNet's data (113GB). Thereafter, with the induction of RoBERTa [15], a large BERT-based model trained on significantly more data than BERT (160GB), outperformed XLNet. Thus, an architecture that is more generalizable and further is trained on larger data, results in NLP benchmarks.

The above-mentioned architectures are primarily language understanding models, where a natural dialect is mapped to a formal interpretation. Here the initial goal is the translation of an input user utterance into a conventional phrase representation. For Natural Language Understanding (NLU) the intermediate representation for the above models' end goal is dictated by the downstream tasks.

Meanwhile, fine-tuning was transpiring to be progressively challenging for task-specific roles in NLU models as it required a greater sample size to learn a particular task, which bereft such models from generalization [16]. This triggered the advent of Natural Language Generation (NLG) models that contrary to NLU training, generated dialect utterances learned from their corresponding masked or corrupted input semantics. Such models operate differently from a routine downstream approach of cursory language comprehension and are optimal for sequence-to-sequence generation tasks, such as language translation. Models like T5 [17], BART [18], mBART [19], T-NLG [20] were pre-trained on a large corpus of corrupted text and generated its corresponding cleaned text via denoising objective [21]. This transition was useful as the additional fine-tuning layer for NLU tasks was not required for NLG purposes. This further enhanced prediction ability via zero or few-shot learning which enabled sequence generation with minimal or no fine-tuning. For instance, if a model's semantic embedding space is pre-trained with animal identification of "cat", "lion" and "chimpanzee", it could still correctly predict "dog" without fine-tuning. Despite superior sequence generation capabilities, NLG model sizes surged exponentially with the subsequent release of GPT-III [22] which was the largest model before the release of GShard [23].

Since NLU and NLG's exceptionally large-sized models required several GPUs to load, this turned out costly and resource prohibitive in most practical situations. Further, when trained for several days or weeks on GPU clusters, these colossal models came at an exorbitant energy cost. To mitigate such computational costs [24], Knowledge Distillation (KD) [25] based models like DistilBERT [26], TinyBERT [27], MobileBERT [28] were introduced at reduced inference cost and size. These smaller student models capitalized on the inductive bias of larger teacher models (BERT) to achieve faster training time. Similarly, pruning and quantization [29] techniques got popular to build economically sized models. Pruning can be classified into 3 categories: weight pruning, layer pruning, and head pruning where certain minimal contributing weights, layers, and attention heads are removed from the model. Like pruning, training-aware quantization is performed to achieve less than 32-bit precision format thereby reducing model size.

For higher performance, greater learning was required which resulted in larger data storage and model size. Due to the model's enormity and implicit knowledge storage, its learning ability had caveats in terms of efficient information access. Current Knowledge Retrieval models like ORQA [30], REALM [31], RAG [32], DPR [33] attempt to alleviate implicit storage concerns of language models by providing external access to interpretable modular knowledge. This was achieved by supplementing the language model's pre-training with a 'knowledge retriever' that facilitated the model to effectively retrieve and attend over explicit target documents from a large corpus like Wikipedia.

Further, the Transformer model's inability to handle input sequences beyond a fixed token span inhibited them to comprehend large textual bodies holistically. This was particularly evident when related words were farther apart than the input length. Hence, to enhance contextual understanding, architectures like Transformer-XL [34], Longformer [35], ETC [36], Big Bird [37], were introduced with modified attention mechanisms to process longer sequences.

Also, due to the surge in demand for NLP models to be economically viable and readily available on edge devices, innovative compressed models were launched based on generic techniques. These are apart from the Distillation, Pruning, and Quantization techniques described earlier. Such models deploy a wide range of computing optimization procedures ranging from hashing [38], sparse attention [39], factorized embedding parameterization [40], replaced token detection [41], inter-layer parameter sharing [42], or a combination of the above mentioned.

## II. RELATED REVIEWS/TAXONOMY

We propose a novel NLP based taxonomy providing a unique classification of current NLP models from six different perspectives:

- *NLU Models*: NLU models excel in classification, structured prediction, and/or query generation tasks. This is accomplished through pre-training and fine-tuning motivated by the downstream task.
- *NLG Models*: Contrary to NLU models, these stand out in sequence-to-sequence generation tasks. They generate clean text via few and single-shot learning from corresponding corrupted utterances.
- *Model Size Reduction*: Use compression-based techniques like KD, Pruning, and Quantization to make large models economical and pragmatic. It's useful for the real-time deployment of large language models to operate on edge devices.
- *Information Retrieval (IR)*: Contextual open domain question answering (QA) is reliant on effective and efficient document retrieval. Hence, IR systems via superior lexical and semantical extraction of physical documents from a large textual corpus create SOTA in the QA domain on multiple benchmarks outperforming contemporary language models.

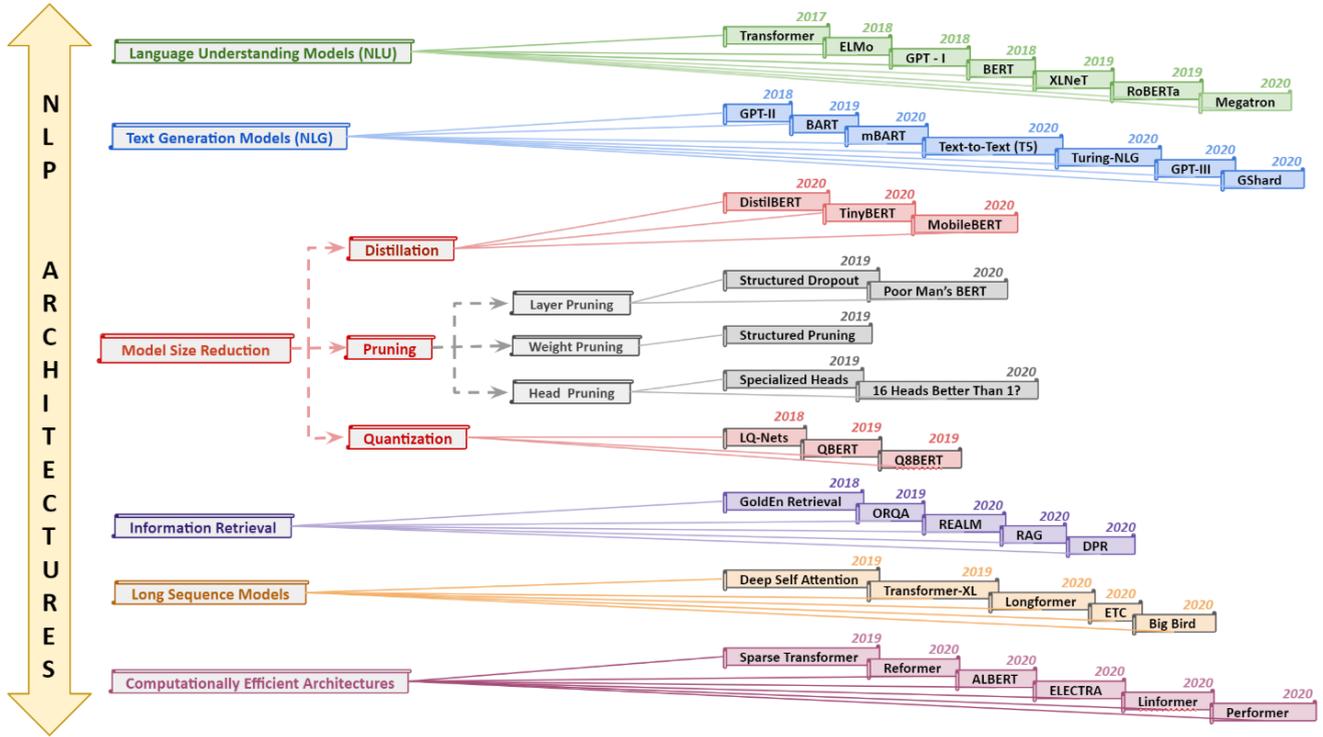

**Figure 1**: Taxonomy of NLP Architectures

- *Long Sequence Models*: Attention-based computational complexity in Transformers scales quadratically with input length, hence it is usually fixed to 512 tokens. This might be acceptable for co-reference resolution tasks that benefit from smaller input lengths [43], however, is inadequate for Question Answering (QA) tasks where reasoning is required across multiple lengthy documents e.g., the HotpotQA dataset [44].
- *Computationally Efficient Architectures*: Memory efficient architectures with comparable accuracies to large language models were built to reduce the high training time of such models.

The above mentioned is a generalized categorization and not a hard classification, few models can be used interchangeably that might serve dual purposes, however, there is a clear demarcation despite insignificant universality. Figure 1 depicts this taxonomy giving a visual breakdown of the significant models belonging to different categories along with their launch years.

### III. PRELIMANIRES TO MODERN NLP ARCHITECTURES

A traditional RNN Encoder-Decoder model [45] comprises of two recurrent neural networks (RNN), where one produces the encoded version of the input sequence, and the other generates its decoded version into a different sequence. To maximize the target's conditional probability for an input sequence, the model is trained jointly with the following language modeling,

$$y^* = argmax \, P(y_t \mid y_1, y_2, y_3, \ldots, y_{t-1}) \quad (1)$$
$$P(y_t \mid y_1, y_2, y_3, \ldots, y_{t-1}) = P(y_t \mid y_1^{t-1}) \quad (2)$$

Such a system is empirically found to give superior results than vanilla RNNs, LSTMs [46], or GRUs [47] by implementing conditional probabilities of phase pairs in machine translation, sequence to sequence mapping, or text summarization tasks.

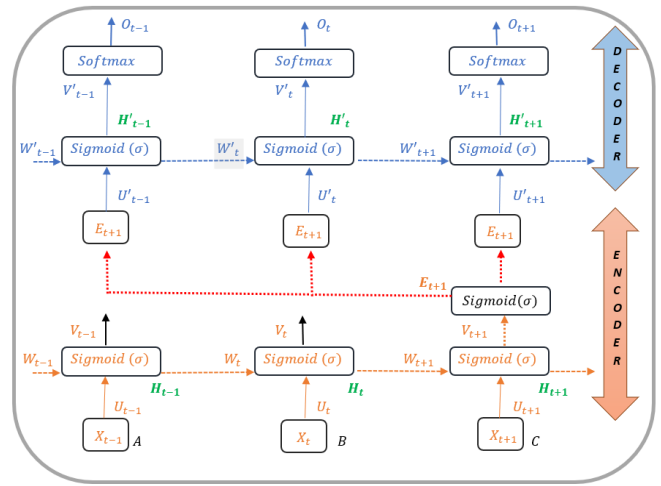

**FIGURE 2.** Encoder-Decoder Architecture

In the above architecture (Figure 2), Encoder's final layer $E_{t+1}$ transmits information to the decoder from its final hidden $V_{t+1}$ the layer which contains the entire contextual understanding of all previous words via a probability distribution.

This combined abstract representation of all the words is fed to the decoder to compute the desired language-based task. Like its preceding layers, the final layer's corresponding learnable parameters are $U_{t+1}$ and $V_{t+1}$ at input and output respectively at the Encoder and $U'_{t+1}, V'_{t+1}$ at the Decoder. Combining the weight matrices with hidden state and bias can be expressed mathematically as follows:

Encoder:
$$H_{t+1} = \sigma(U_{t+1}.X_{t+1} + W_{t+1}.H_t + b_t) \quad (3)$$
$$E_{t+1} = \sigma(V_{t+1}.H_{t+1} + b_t) \quad (4)$$
Decoder:
$$H'_{t+1} = \sigma(U'_{t+1}.E_{t+1} + W'_{t+1}.H'_t + b_{t+1}) \quad (5)$$
$$O_{t+1} = Softmax(H'_{t+1}.V'_{t+1} + b_{t+1}) \quad (6)$$

Thereafter, the induction of Attention [48],[49] in 2014-15 overcame the RNN Encoder-Decoder limitation that suffered from prior input dependencies, making it challenging to infer longer sequences and suffered from vanishing and exploding gradients [50]. The attention mechanism eliminated the RNN dependency by disabling the entire input context through one final Encoder node. It weighs all inputs individually that feed the decoder to create the target sequence. This results in a greater contextual understanding leading to superior predictions in target sequence generation. First, the alignment determines the extent of match between the $j^{th}$ input and $i^{th}$ output which can be determined as
$$e_{tj} = \tanh(h_{i-1}, h_j) \quad (7)$$
More precisely, the alignment scores take as input all encoder output states and the previous decoded hidden state which is expressed as:
$$Score_{Align} = W_{comb}.\tanh(W_{dec}.H_{dec} + W_{enc}.H_{enc}) \quad (8)$$

The decoder's hidden state and encoder outputs are passed via their respective linear layers along with their trainable weights. The weight $\alpha_{tj}$ for each encoded hidden representation $h_j$ is computed as:
$$\alpha_{tj} = \frac{\exp(e_{tj})}{\sum_{k=1}^{T_x} \exp(e_{tk})}, \quad (9)$$
The resulting context vector in this attention mechanism is determined by:
$$c_t = \sum_{j=1}^{T_x} \alpha_{tj} h_j \text{ where } T_x = input\ sequence\ length \quad (10)$$
The Attention mechanism is essentially the generation of the context vector computed from the various alignment scores at different positions as shown in figure 3.
Luong's Attention mechanism differs from the above-mentioned Bahdanau's in terms of alignment score computation. It uses both global and local attention, where the global attention uses all encoder output states while the local attention focuses on a small subset of words. This helps to achieve superior translation for lengthier sequences. These attention designs led to the development of modern Transformer architectures which use an enhanced attention mechanism as described in the next section.

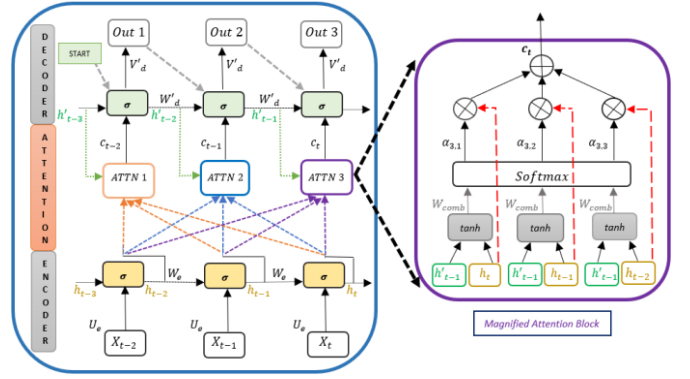

**FIGURE 3.** Attention Mechanism on Encoder-Decoder Model

## IV. NLU ARCHITECTURES
NLU's approach of transferring pre-trained neural language representations demonstrated that pre-trained embeddings improve downstream task results when compared to embeddings learned from scratch [51],[52]. Subsequent research works enhanced learning to capture contextualized word representations and transferred them to neural models [53],[54]. Recent efforts not limited to [55],[56],[57] have further built on these ideas by adding end-to-end fine-tuning of language models for downstream tasks in addition to extraction of contextual word representations. This engineering progression, coupled with large compute availability has evolved NLU's state of the art methodology from transferring word embeddings to transferring entire multi-billion parameter language models, achieving unprecedented results across NLP tasks. Contemporary NLU models leverage Transformers for modeling tasks and exclusively use an Encoder or a Decoder-based approach as per requirements. Such models are vividly explained in the subsequent section.

### IV-A TRANSFORMERS
*IV-A.1. The Architecture*
The original Transformer is a 6-layered Encoder-Decoder model, that generates a target sequence via the Decoder from the source sequence via the Encoder. The Encoder and Decoder at a high level consist of a self-attention and a feed-forward layer. In the Decoder an additional attention layer in between enables it to map its relevant tokens to the Encoder for translation purposes. Self Attention enables the look-up of remaining input words at various positions to determine the relevance of the currently processed word. This is performed for all input words that help to achieve a superior encoding and contextual understanding of all words.
Transformer architecture was built to induct parallelism in RNN and LSTM's sequential data where input tokens are fed instantaneously and corresponding embeddings are generated simultaneously via the Encoder. This embedding maps a word (token) to a vector that can be pre-trained on the fly, or to conserve time a pre-trained embedding space like GloVe is implemented. However, similar tokens in different sequences might have different interpretations which are resolved via a positional encoder that generates context-based word information concerning its position. Thereafter the enhanced contextual representation is fed to the attention layer which furthers contextualization by generating attention vectors, that determine the relevance of the $i^{th}$ word in a sequence concerning other words. These attention vectors are then fed to the feed-forward Neural Network where they are transformed to a more digestible form for the next 'Encoder' or Decoder's 'Encoder-Decoder Attention' block.

The latter is fed with Encoder output and Decoder input embedding that performs attention between the two. This determines the relevance of Transformer's input tokens concerning its target tokens as the decoder establishes actual vector representation between the source and target mapping. The decoder predicts the next word via softmax which is executed over multiple time steps until the end of the sentence token is generated. At each Transformer layer, there are residual connections followed by a layer normalization [58] step to speed up the training during backpropagation. All of the transformer architectural details are demonstrated in Figure 4.

### IV-A.2. Queries, Keys, and Values

The input to the Transformer's Attention mechanism is target token Query vector $Q$, its corresponding source token Key vector $K$, and Values $V$ which are embedding matrices. Mapping of source and destination tokens in machine translation can be quantified as to how similar each of their tokens is in a sequence via inner dot product. Therefore, to achieve accurate translation the key should match its corresponding query, via a high dot product value between the two. Assume $Q \in \{L_Q, D\}$ and $K \in \{L_K, D\}$ where $L_Q, L_K$ represent target and source lengths, while $D$ denotes the word embedding dimensionality. Softmax is implemented to achieve a probability distribution where all Query, Key similarities add up to one and make attention more focused on the best-matched keys.

$$W_{SM} = softmax(Q.K^T) \text{ where } W_{SM} \in \{L_Q, L_K\} \quad (11)$$

Query assigns a probability to key for matching and often values are similar to keys, therefore
$$Z_{Att} = Attention(Q, K, V) = softmax(Q.K^T).V = W_{SM}.V \quad (12)$$

### IV-A.3. Multi-Headed Attention (MHA) and Masking

MHA enhances the model's capacity to emphasize a sequence's different token positions by implementing attention parallelly multiple times. The resulting individual attention outputs or heads are concatenated and transformed via a linear layer to the expected dimensions. Each of the multiple heads enables attending the sequence parts from a different perspective providing similar representational forms for each token. This is performed as each token's self-attention vector might weigh the word it represents higher than others due to the high resultant dot product. This is not productive since the goal is to achieve similarly assessed interaction with all tokens. Therefore self-attention is computed 8 different times resulting in 8 separate attention vectors for each token which are used to compute the final attention vector via a weighted sum of all 8 vectors for each token. The resultant multi-headed attention vectors are computed in parallel which is fed to the feed-forward layer. Each subsequent target token $T_{t+1}$ is generated using as many source tokens in the encoder $(S_0, .., S_{t+n})$. However, in an autoregressive decoder only previous time stepped target tokens are considered $(T_0, .., T_t)$, for future target prediction purposes known as causal masking. This is provided to enable maximal learning of the subsequently translated target tokens. Therefore during parallelization via matrix operations, it is ensured that the subsequent target words are masked to zero, so the attention network cannot see into the future. The Transformer described above resulted in significant improvement in the NLP domain. This leads to a plethora of high-performance architectures that we describe in the subsequent sections.

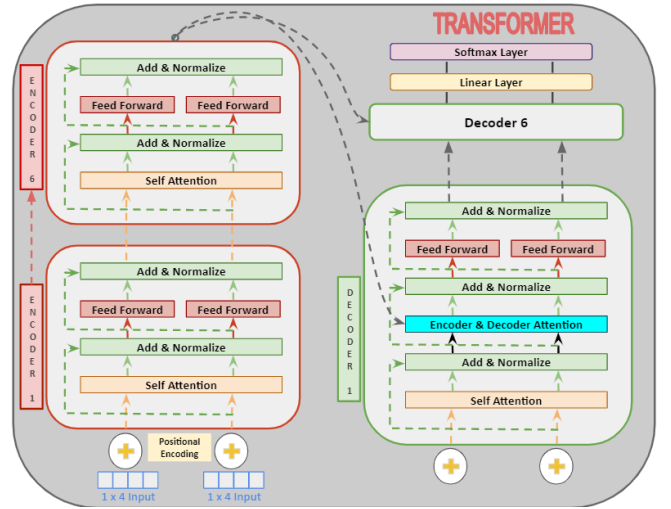

**FIGURE 4.** The Multi-headed Transformer Architecture

### IV-B EMBEDDINGS FROM LANGUAGE MODELS: ELMo

The goal of ELMo [59] is to generate a deep contextualized word representation that could model (i) intricate syntactical and semantical characteristics of word (ii) polysemy or lexical ambiguity, words with similar pronunciations could have different meanings at different contexts or locations. These enhancements gave rise to contextually rich word embeddings which were unavailable in the previous SOTA models like GloVe. Unlike its predecessors that used a predetermined embedding, ELMo considers all $N$ token occurrences $(t_1, t_2, .., t_N)$ for each token $t$ in the entire sequence before creating embeddings. The authors hypothesize that the model could extract abstract linguistic attributes in its architecture's top layers via a task-specific bi-directional LSTM.

This is possible by combining a forward and a backward language model. At timestep $k-1$, the forward language model predicts the next token $t_k$ given the input sequence's previous observed tokens via a joint probability distribution shown in (13). Likewise, in (14) with its order reversed, the backward language model forecasts the prior tokens given the future tokens.

$$p(t_1, t_2, \ldots, t_n) = \prod_{k=1}^{N} p(t_k \mid t_1, t_2, \ldots, t_{k-1}) \qquad (13)$$

$$p(t_1, t_2, \ldots, t_n) = \prod_{k=1}^{N} p(t_k \mid t_{k+1}, t_{k+2}, \ldots, t_N) \qquad (14)$$

This is further implemented through a softmax on top of the final LSTM layer as shown in Figure 5.

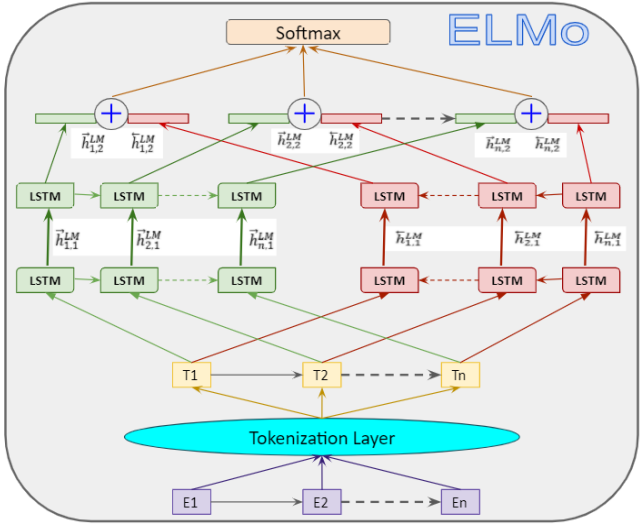

**FIGURE 5.** Bi-directional LSTM based ELMo Language Model

ELMo for each token representation $x_k$ computes its intermediary bi-directional vector representation $h_{k,j}$ at each layer $j$ of the LSTM model as:

$$R_k = \{ x_k^{LM}, \vec{h}_{k,j}^{LM}, \overleftarrow{h}_{k,j}^{LM} \mid j = 1, \ldots, L \}$$
$$= \{ h_{k,j}^{LM} \mid j = 0, \ldots, L \} \qquad (15)$$

Mathematically $h_{k,0}^{LM} = x_k$ will the lowest level token representation and it could be generalized as:

$$h_{k,j}^{LM} = [\vec{h}_{k,j}^{LM}, \overleftarrow{h}_{k,j}^{LM}] \; j \in \{1, \ldots, L\} \qquad (16)$$

ELMo learns normalized weights via softmax $s_j^{task}$ over $L$ layer representations. This results in a task-specific hyperparameter $\gamma^{task}$ that enables the task's scaling optimization. Hence for a particular task, the word representation variance in different layers is expressed as:

$$ELMo_k^{task} = E(R_k; \theta^{task}) = \gamma^{task} \sum_{j=0}^{L} s_j^{task} h_{k,j}^{LM} \qquad (17)$$

### IV-B GENERATIVE PRE-TRAINING MODEL: GPT-I

In the first phase through unsupervised learning, the decoder-based GPT-I is pre-trained on a large dataset. This promotes raw data compute that eliminates the data labeling bottleneck of supervised learning. The second phase performs task-specific fine-tuning on considerably smaller supervised datasets with marginal input variations. Consequently, it led to greater task agnosticism than then SOTA models like ELMo, ULMFiT [60] and succeeded in more sophisticated tasks like common-sense reasoning, semantic similarity, and reading comprehension. The pre-training of GPT-I can be modeled as a maximization function of unsupervised tokens $\{u_i, \ldots, u_n\}$.

$$L_1(\mathcal{U}) = \sum_i \log P(u_i \mid u_{i-k}, \ldots, u_{i-1}; \Theta) \qquad (18)$$

where $k$ is the context window size and conditional probability is parametrized via $\Theta$. With multi-headed-attention and feedforward layers, a target token-based probability distribution via softmax is produced.

$$h_n = transformer_{block(h_{l-1})} \forall i \in [1, n] \qquad (19)$$
$$h_0 = UW_e + W_p \qquad (20)$$
$$P(u) = softmax(h_n W_e^T) \qquad (21)$$

where $(U = u_{-k}, \ldots, u_{-1})$ is the set of context token vector, $n$ is the number of layers, $W_e$ and $W_p$ are the token and positional embedding matrices respectively. Post-pre-training, parameter adaptation for the supervised end task takes place. Here input sequence $(x^1, \ldots, x^m)$ from a labeled dataset $\mathcal{C}$ is fed to the previous pre-trained model to obtain the transformer's block final activation $h_l^m$ that is fed to a parametrized $(W_y)$ linear output layer for prediction $(y)$. Also, the objective $L_2(\mathcal{C})$ is maximized is as follows

$$P(y \mid x^1, \ldots, x^m) = softmax(h_l^m W_y) \qquad (22)$$
$$L_2(\mathcal{C}) = \sum_{(x,y)} \log P(y \mid x^1, \ldots, x^m) \qquad (23)$$

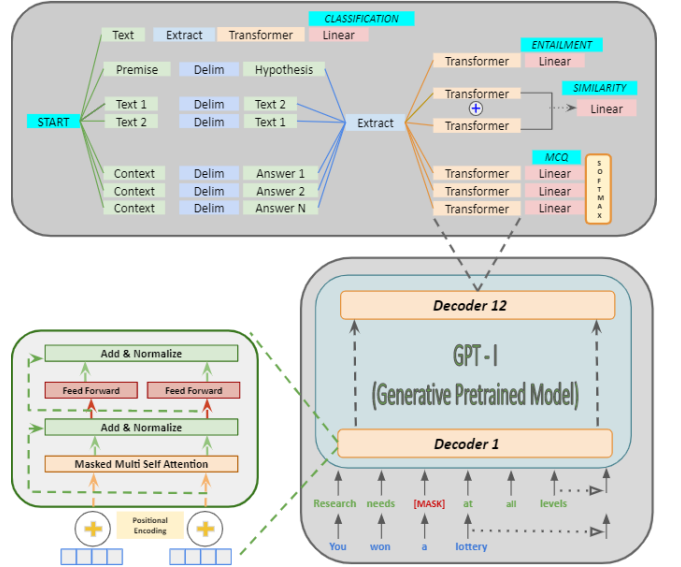

**FIGURE 6.** GPT-1 task-based Architecture (top) and magnified views of Transformer based Decoder (bottom)

Incorporating a secondary language modeling objective during fine-tuning enhances learning by a better generalization of the supervised model and accelerates convergence as:

$$L_3(\mathcal{C}) = L_2(\mathcal{C}) + \lambda L_1(\mathcal{C}) \qquad (24)$$

GPT performs various tasks like classification, entailment, similarity index, Multiple-Choice Questions (MCQ) as shown in figure 6. The extraction phase distills features from textual bodies before which the text is separated via the 'Delimiter' token during text pre-processing. This token is not required for classification tasks since it does not need to gauge the relationship between multiple sequences. Moreover, Q&A or textual entailment tasks involve defined inputs like ordered sentence pairs or triplets in a document. For MCQ tasks, contextual alterations are required at input to achieve the correct results. This is done via a Transformer based Decoder training objective where input transformations are fine-tuned for their respective answers.

*IV-C BIDIRECTIONAL ENCODER REPRESENTATIONS FROM TRANSFORMER: BERT*

BERT is a stack of pre-trained Transformer Encoders that overcomes prior models' restrictive expressiveness i.e., GPT's lack of bidirectional context and ELMo's shallow dual context's concatenation. BERT's deeper model provides a token with several contexts with its multiple layers and the bi-directional model provides a richer learning environment. However, bi-directionality raises concerns that tokens could implicitly foresee future tokens during pre-training resulting in minimal learning and leading to trivial predictions. To effectively train such a model, BERT implements Masked Language Modeling (MLM) that masks 15% of all input tokens randomly in each input sequence. This masked word prediction is the new requirement unlike recreating the entire output sequence in a unidirectional LM. BERT masks during pre-training, hence the [MASK] token does not show during fine-tuning, creating a mismatch as the "masked" tokens are not replaced. To overcome this disparity, subtle modeling modifications are performed during the pre-training phase. If a token $T_i$ is chosen to be masked, then 80% of the time it is replaced with the [MASK] token, 10% of the time a random token is chosen and for the remaining 10%, it remains unchanged. Thereafter $T_i$ cross-entropy loss will predict the original token, the unchanged token step is employed to maintain a bias towards the correct prediction. This methodology creates a state of randomness and constant learning for the Transformer encoder which is compelled to maintain a distributed contextual representation of each token. Further, as random replacement arises for a mere 1.5% of all tokens (10% of 15%), this does not seem to impair the language model's understanding ability.

Language modeling could not explicitly comprehend the association between multiple sequences; therefore it was deemed sub-optimal for inference and Q&A tasks. To overcome this, BERT was pre-trained with a monolingual corpus for a binarized Next Sentence Prediction (NSP) task. As shown in Figure 7, sentences Y (He came [MASK] from home) and Z (Earth [MASK] around Sun) do not form any continuity or relationship. Since Z is not the actual next sentence following Y, the output classification label [NotNext] gets activated, and [IsNext] activates when sequences are coherent.

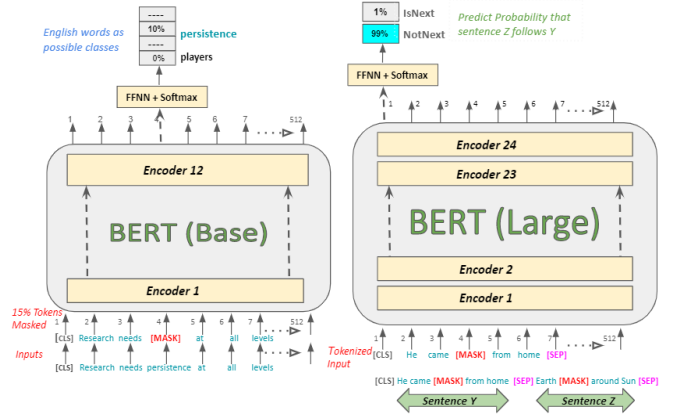

**FIGURE 7.** The architecture of BERT's MLM and NSP functionality

*IV-D GENERALIZED AUTOREGRESSIVE PRETRAINING FOR LANGUAGE UNDERSTANDING: XLNeT*

XLNet captures the best of both worlds where it preserves the benefits of Auto-Regressive (AR) modeling and bidirectional contextual capture. To better comprehend why XLNet outperforms BERT, consider the 5-token sequence [San, Francisco, is, a, city]. The two tokens chosen for prediction are [San, Francisco], hence BERT and XLNet maximize $log\ p(San\ Francisco\ |\ is\ a\ city)$ as follows:

$\mathcal{L}_{BERT} = \log p\ (San|\ is\ a\ city)\ +$
$\qquad \log p\ (Francisco|is\ a\ city)$
$\mathcal{L}_{XLNet} = \log p\ (San|\ is\ a\ city)\ +$
$\qquad \log p\ (Francisco|San\ is\ a\ city)$

The above can further be generalized for the target ($\mathcal{T}$) and non-target token set ($\mathcal{N}$), BERT and XLNet will maximize $\log p\ (\mathcal{T}|\mathcal{N})$ with the following different interpretability:

$$\mathcal{L}_{BERT} = \sum_{x \in \mathcal{T}} \log p(x|\ \mathcal{N}) \qquad (25)$$

$$\mathcal{L}_{BERT} = \sum_{x \in \mathcal{T}} \log p(x|\ \mathcal{N}\mathcal{T}_{<x}) \qquad (26)$$

XLNet considers the target as well as the remaining tokens for prediction, whereas BERT only considers the non-target tokens. Hence, XLNet captures the inter-pair dependency [San, Francisco] unlike BERT where either [San] or [Francisco] leads to correct prediction. Further, via AR XLNet performs factorized ordering on all possible token permutations ($L! = 5!$) of sequence length $L$ in the set i.e., $\{[1, 2, 3, 4, 5], [1, 2, 5, 4, 3], \ldots, [5, 4, 3, 2, 1]\} \cong$ [is, San, Francisco, a, city] etc.

$$\max_{\theta} \quad E_{z \sim \mathcal{Z}_T} \left[ \sum_{t=1}^{T} \log p_\theta(x_{z_t}\ |\ x_{z_{<t}}) \right] \qquad (27)$$

where set $\mathcal{Z}_T$ contains all permutational sequences of length $T\ [1, 2, \ldots, T]$ and $x_{z_t}$ is the reference token. Hence the target learns from numerous combinations attaining a richer contextualized learning. Further for all permutable factorization orders, the model parameters are shared to build knowledge and bidirectional context from all factorizations as demonstrated via equation 27.

## IV-D.1. Masking

There is a challenge to determine the word order in the sequence as the token ($x_{z_t}$) determining the autoregression is not considered. This word order is partially achieved via positional encoding, however, for contextual understanding XLNet employs masking. Consider a generated permutation of [2, 1, 3] in a 3-token sequence where the first token i.e., 2 has no context hence all masking results in [0,0,0] in the 2nd row of the 3×3 masking matrix. Similarly, the 2nd and 3rd masks would result in [0,1,0] and [1,1,0] in the 1st and 3rd row of the Query Stream (QS) masking matrix where the token cannot see itself. QS matrix with an all-one diagonal inclusion constitutes Content Stream (CS) masking matrix where each token can see itself. This 3-token sequence masking is demonstrated in figure 8 below.

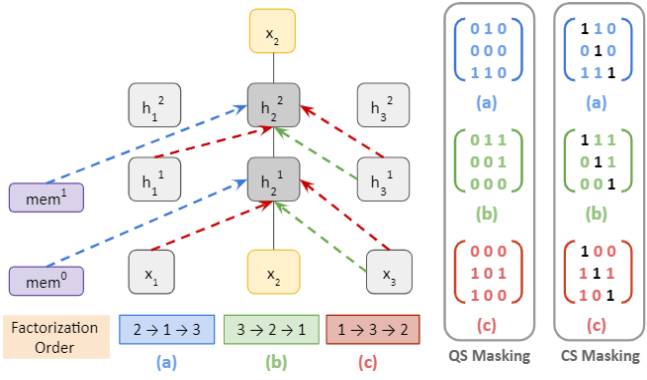

**FIGURE 8.** Illustration of predicting $x_2$ in the 3-token sequence with different factorization orders and its corresponding masking matrices

The first reference '2' has no context which is gathered from its corresponding 'mem block', a Transformer-XL-based extended cached memory access. Thereafter it receives context from token '3' and '1','3' for subsequent orderings.

## IV-D.2. Model Architecture

Figure 9 demonstrates the model's two-stream attention framework that consists of a content and query stream attention process to achieve greater understanding via contextualization. This process is initiated via target-aware representation, where the target position is baked into the input for subsequent token generation purposes.

*(i) Target Aware Representation:* A vanilla implementation of Transformer based parametrization does not suffice for complex permutation-based language modeling. This is because the next token distribution $p_\theta(X_{Z_t} \mid x_{z<t})$ is independent of the target position i.e., $Z_t$. Subsequently, redundant distribution is generated, which is unable to discover effective representations, hence target position-aware re-parametrization for the next-token distribution is proposed as follows:

$$p_\theta(X_{Z_t} = x \mid \mathbf{x}_{z<t}) = \frac{\exp(e(x)^T h_\theta(\mathbf{x}_{z<t}))}{\sum_{x'} \exp(e(x')^T h_\theta(\mathbf{x}_{z<t}))} \quad (28)$$

$$p_\theta(X_{Z_t} = x \mid \mathbf{x}_{z<t}) = \frac{\exp(e(x)^T g_\theta(\mathbf{x}_{z<t}, Z_t))}{\sum_{x'} \exp(e(x')^T g_\theta(\mathbf{x}_{z<t}, Z_t))} \quad (29)$$

where $g_\theta(x_{z<t}, Z_t)$ is a modified representation that additionally considers the target position $Z_t$ as an input.

*(ii) Two Stream Self Attention:* The formulation of $g_\theta$ remains a challenge despite the above resolution as the goal is to rely on the target position $Z_t$ to gather contextual information $x_{z<t}$ via attention, hence: (1) For $g_\theta$ to predict $x_{Z_t}$, it should utilize the position of $Z_t$ only to incorporate greater learning, not the content $x_{Z_t}$ (2) To predict other tokens $x_{Z_j}$ where $j > t$, $g_\theta$ should encode the context $x_{Z_t}$ to provide full contextual understanding.

To further resolve the above conflict, the authors propose two sets of hidden representation instead as follows:

- The hidden content representation $h_\theta(x_{z<t}) \cong h_{Z_t}$ that encodes both context and content $x_{Z_t}$
- The query representation $g_\theta(x_{z<t}, Z_t) \cong g_{Z_t}$ which solely accesses the contextual information $x_{z<t}$ and position $Z_t$ without the content $x_{Z_t}$

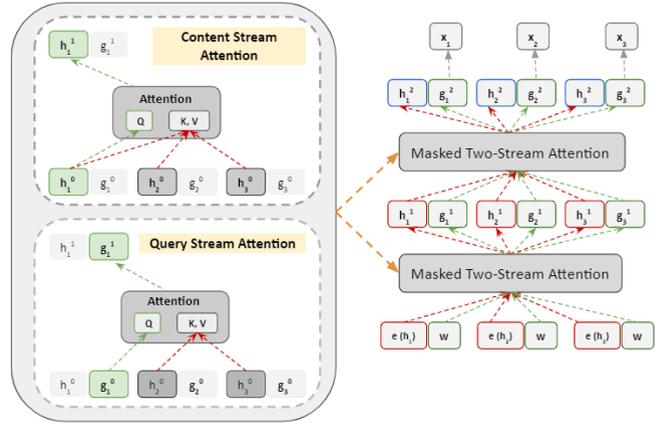

**FIGURE 9.** (Left): Standard Attention via Content Stream and Query Stream Attention without access to the content. (Right): LM training

The above two attention courses are parametrically shared and updated for every self-attention layer $m$ as:

$$Attention(Q = h_{Z_t}^{(m-1)}, KV = h_{Z_{\leq t}}^{(m-1)}; \theta) \to h_{Z_t}^{(m)}$$
(Content Stream: utilize both $Z_t$ and $x_{Z_t}$)

$$Attention(Q = g_{Z_t}^{(m-1)}, KV = h_{Z_{<t}}^{(m-1)}; \theta) \to g_{Z_t}^{(m)}$$
(Query Stream: use $Z_t$ without seeing $x_{Z_t}$)

This dual attention is pictorially expressed in figure 9. For simplicity purposes, consider the prediction of token $t_i$ that is not allowed to access its corresponding embedding from the preceding layer. However, to predict $t_{i+1}$ the token $t_i$ needs to access its embedding and both operations must occur in a single pass.

Therefore, two hidden representations are implemented where $h_{Z_t}^{(m)}$ is initialized via token embeddings and $g_{Z_t}^{(m)}$ through weighted transformations. From above equations $h_{Z_t}^{(m)}$ can access the history including the current position whereas $g_{Z_t}^{(m)}$ can access only previous $h_{Z_t}^{(m)}$ positions. The token prediction happens in the final layer via $g_{Z_t}^{(m)}$. For greater sequence length processing the memory blocks are derived from Transformer-XL which can process longer than standard Transformer input sequence lengths. The hidden representations mentioned above are also stored in the memory blocks.

### IV-E A Robustly Optimized BERT Pretraining Approach: RoBERTa

This paper claimed that BERT was considerably undertrained and as a result, RoBERTa incorporated a greater training intensive regime. This was for BERT-based models that could match or exceed the prior methods. Their revisions include: *(i)* longer training duration with greater data and batch sizes *(ii)* eliminating BERT's NSP goal *(iii)* longer sequence training *(iv)* training data's masking pattern modified dynamically. The authors claim superior performance over BERT on downstream tasks for a more diverse and voluminous CC-News dataset.

Further, BERT implements a non-efficient static masking implementation to avoid redundant masks. For instance, training data that is duplicated 10 times for a sequence to be masked in 10 different ways for 40 training epochs, where each training sequence is seen with the same mask 4 times. RoBERTa provides slightly enhanced results via incorporating dynamic masking where a masking pattern is generated each time the model is fed a sequence while pretraining larger datasets. Recent work has questioned BERT's NSP [61] role which was conjectured to play a key role in its performance in language inference and Q&A tasks. RoBERTa amalgamates both hypotheses and provides numerous supplementary training formats that perform like BERT and outperform it for full sentence training excluding the NSP loss. RoBERTa provides similar and marginally better results than BERT on GLUE benchmark as well as on RACE and SQUAD datasets without fine-tuning for multiple tasks.

### IV-E MEGATRON LANGUAGE MODEL (LM)

Megatron was the largest model when released with the size of $24 \times$ BERT and $5.6 \times$ GPT-2 and could not fit in a single GPU. Hence the key engineering implementation was the induction of its 8 and 64-way model, and data parallelized version where parameters were split across (~512) GPUs. It sustained high performance (15.1 Petaflops) and scaling efficiency (76%), whereas BERT resulted in performance degradation with size growth. This feat was primarily attributed to layer normalization and residual connection re-ordering within the transformer layers. This led to monotonically superior performance on downstream tasks with increased model size.

Megatron overcomes the prior model's memory constraint via splitting the model across several accelerators. This not only resolves the memory usage but enhances the model parallelism irrespective of batch size. It incorporates distributed tensor computations to upsurge model size or acceleration and parallelizes the attention head computation. This does not require a new compiler or code re-write and is implementable with a few parameters.

First, the Multi-Layer Perceptron (MLP) block partitions the GEMM parallelly in two columns, enabling GeLU nonlinearity applied independently to each partitioned GEMM. This GeLU output is fed directly to the row-wise parallelized GEMM whose output is reduced via a single all-reduce operator (*g and f*) in forward and backward pass before passing it to the dropout layer.

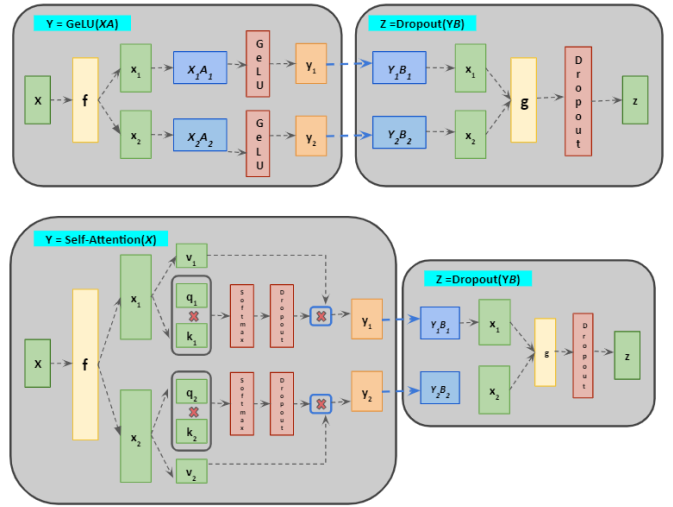

**FIGURE 10.** Parallelized Megatron's MLP and Self-Attention blocks

Parallelism in the self-attention block is achieved by partitioning the GEMMs column-wise for each key, query, and value set. Hence, the workload is split across all GPUs as matrix multiplication for each attention head is performed on a single GPU. The resulting GEMM output, like MLP undergoes an all-reduce operation and is parallelized across rows as shown above in figure 10. This technique eliminates the need for synchronization between GEMMs for MLP and attention blocks.

## V. NLG ARCHITECTURES

In NLU models, the sheer amount of data compute required for learning numerous post pre-trained 'fine-tuned' tasks is parametrically inefficient, as an entirely new model is required for every task. These models can be exemplified as narrow experts rather than proficient generalists. Therefore, NLG models provide a transition towards building generic systems, that accomplish several tasks without the necessity to create and label a training dataset manually for each task. Moreover, MLM in NLU models is unable to capture a rich relationship between multiple sequences. Further, most effective NLU models derive their methodologies from the MLM model variants which are denoising autoencoders trained on text reconstruction where a random subset of words is masked out. Consequently, NLG models in the last few years have made tremendous progress on tasks like text translation and summarization, Q&A, NLI, conversational engagement, picture description, with unprecedented accuracies.

## V-A LANGUAGE MODELS ARE UNSUPERVISED MULTI-TASK LEARNERS: GPT-II

GPT-II [62] was possibly the first model that dawned on the rise of NLG models. It was trained in an unsupervised manner capable of learning complex tasks including Machine Translation, reading comprehension, and summarization without explicit fine-tuning. Task-specific training corresponding to its dataset was the core reason behind the generalization deficiency witnessed in current models. Hence robust models would likely require training and performance gauges on a variety of task domains.

GPT-II incorporates a generic probabilistic model where numerous tasks can be performed for the same input as $p(output|input, task)$. The training and test set performance improves as model size is scaled up and as a result, it under fits on the huge WebText dataset. The 1.5 billion parameter GPT-2 outperformed its predecessors on most datasets in the previously mentioned tasks in a zero-shot environment. It is an extension of the GPT-I decoder-only architecture trained on significantly greater data.

## V-B BIDIRECTIONAL AND AUTOREGRESSIVE TRANSFORMERS: BART

A denoising autoencoder BART is a sequence-to-sequence [63] model that incorporates two-stage pre-training: (1) Corruption of original text via a random noising function, and (2) Recreation of the text via training the model. Noising flexibility is the major benefit of the model where random transformations not limited to length alterations are applied to the original text. Two such noising variations that stand out are random order shuffling of the original sentence and a filling scheme where texts of any spanned length are randomly replaced by a single masked token. BART deploys all possible document corruption schemes as shown below in figure 11, wherein the severest circumstance all source information is lost and BART behaves like a language model.

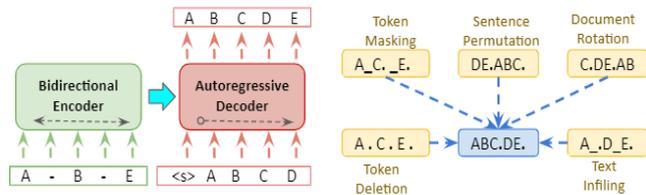

FIGURE 11. Denoised BART Model and its Noising Schemes

This forces the model to develop greater reasoning across overall sequence length enabling greater input transformations which results in superior generalization than BERT. BART is pre-trained via optimization of a reconstruction loss performed on corrupted input documents i.e., cross-entropy between decoder's output and original document. For machine translation tasks, BART's encoder embedding layer is replaced with an arbitrarily initialized encoder, that is trained end-to-end with the pre-trained model as shown in Figure 12. This encoder maps its foreign vocabulary to BART's input which is denoised to its target language English. The source encoder is trained in two stages, that share the backpropagation of cross-entropy loss from BART's output. Firstly, most BART parameters are frozen, and only the arbitrarily initialized encoder, BART's positional embeddings, and its encoder's self-attention input projection matrix are updated. Secondly, all model parameters are jointly trained for few iterations. BART achieves state-of-the-art performance on several text generation tasks, fueling further exploration of NLG models. It achieves comparative results on discriminative tasks when compared with RoBERTa.

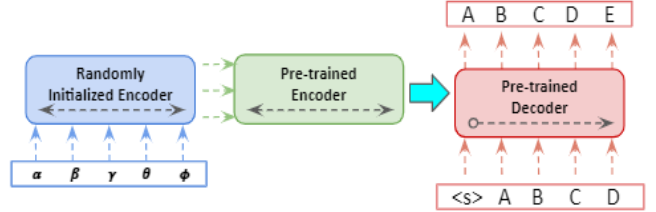

FIGURE 12. Denoised BART Model for fine-tuned MT tasks

## V-C MULTILINGUAL DENOISING PRE-TRAINING FOR NEURAL MACHINE TRANSLATION: mBART

### V-C.1. Supervised Machine Translation

mBART demonstrates that considerable performance gains are achieved over prior techniques [64], [65] by autoregressively pre-training BART, via sequence reconstructed denoising objective across 25 languages from the common crawl (CC-25) corpus [66]. mBART's parametric fine-tuning can be supervised or unsupervised, for any linguistic pair without task-specific revision. For instance, fine-tuning a language pair i.e. (German-English) enables the model to translate from any language in the monolingual pre-training set i.e. (French English), without further training. Since each language contains tokens that possess significant numerical variations, the corpus is balanced via textual up/downsampling from each language $i$ with the ratio $\lambda_i$

$$\lambda_i = \frac{1}{p_i} \cdot \frac{p_i^{\alpha}}{\sum_i p_i^{\alpha}} \qquad (30)$$

where $p_i$ is each language's percentage in the dataset with a soothing parameter $\alpha = 0.7$. The training data encompasses $K$ languages: $\mathcal{C} = \{\mathcal{C}_1, --, \mathcal{C}_k\}$ where each $\mathcal{C}_i$ is $i^{th}$ language's monolingual document collection. Consider a text corrupting noising function $g(X)$ where the model is trained to predict original text $X$, hence loss $\mathcal{L}_\theta$ is maximized as:

$$\mathcal{L}_\theta = \sum_{\mathcal{C}_i \in \mathcal{C}} \sum_{X \in \mathcal{C}_i} \log P(X \mid g(X); \theta) \qquad (31)$$

where language $i$ has an instance $X$ and above distribution $P$ is defined via a sequence-to-sequence model.

### V-C.2. Unsupervised Machine Translation

mBART is evaluated on tasks where target bi-text or text pairs are not available in these 3 different formats.
- ❖ None of any kind of bi-text is made available, here back-translation (BT) [67],[68] is a familiar solution. mBART

- offers a clean and effective initialization scheme for such techniques.
- ❖ The bi-text for the target's pair is made unavailable, however, the pair is available in the target language's bi-text corpora for other language pairs.
- ❖ Bi text is not available for the target pair, however, is available for translation from a different language to the target language. This novel evaluation scheme demonstrates mBART's transfer learning capability despite the absence of the source language's bi-text mBART is pre-trained for all 25 languages and fine-tuned for the target language as shown in figure 13.

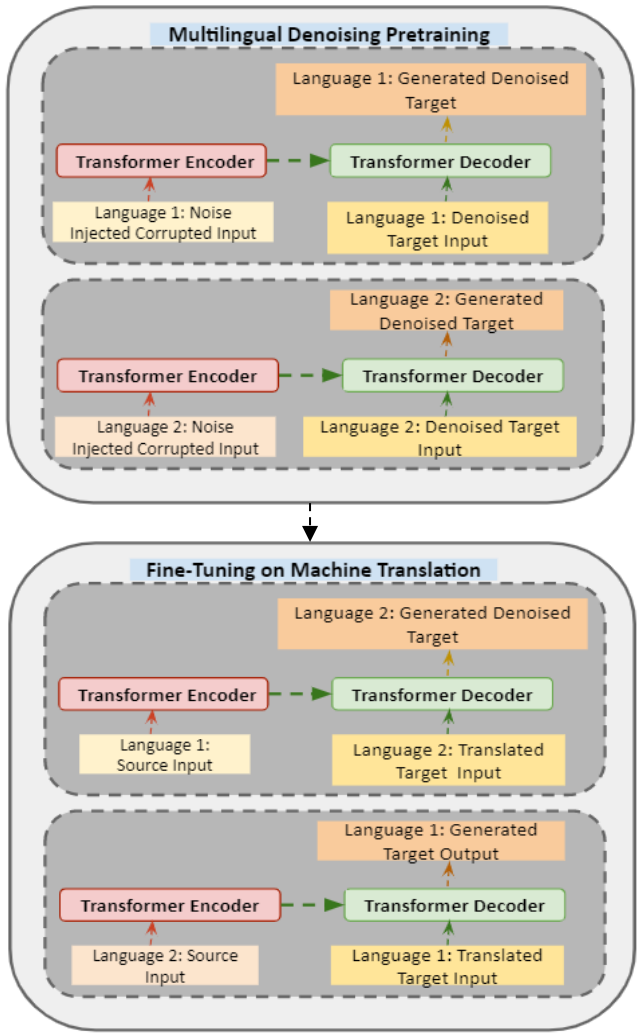

**FIGURE 13.** mBART Generative Model Pre-training & Fine-tuning

### V-D EXPLORING THE LIMITS OF TRANSFER LEARNING WITH A TEXT-TO-TEXT TRANSFORMER: T5

This model was built by surveying and applying the most effective transfer learning practices. Here all NLP tasks are orchestrated within the same model and hyperparameters are reframed into a unified text-to-text setup where text strings are inputs and outputs. A high-quality, diverse and vast dataset is required to measure the scaled-up effect of pre-training in the 11 billion parameter T5. Therefore, Colossal Clean Crawled Corpus (C4) was developed, twice as large as Wikipedia.

The authors concluded that causal masking limits the model's capability to attend only till the $i^{th}$ input entry of a sequence, which turns detrimental. Hence T5 incorporates fully visible masking during the sequence's prefix section (prefix LM) whereas causal masking is incorporated for training the target's prediction. The following conclusions were made after surveying the current transfer learning landscape.

- ❖ Model Configuration: Normally models with Encoder-Decoder architectures outperformed decoder-based language models.
- ❖ Pre-Training Goals: Denoising worked best for fill-in-the-blank roles where the model is pre-trained to retrieve input missing words at an acceptable computational cost
- ❖ In-Domain Datasets: In-domain data training turns out to be effective, however pre-training small datasets generally leads to overfitting.
- ❖ Training Approaches: A pre-train, fine-tune methodology for multi-task learning could be effective, however, each task's training frequency needs to be monitored.
- ❖ Scaling Economically: To efficiently access the finite computing resources, evaluation among model size scaling, training time, and ensembled model quantity is performed.

### V-E TURING NATURAL LANGUAGE GENERATION: T-NLG

T-NLG is a 78 layered Transformer based generative language model, that outsizes the T5 with its 17 billion trainable parameters. It possesses greater speedup than Nvidia's Megatron, which was based on interconnecting multiple machines via low latency buses. T-NLG is a progressively larger model, pre-trained with greater variety and quantity of data. It provides superior results in generalized downstream tasks with lesser fine-tuning samples. Hence, its authors conceptualized training a huge centralized multi-task model with its resources shared across various tasks, rather than allocating each model for a task. Consequently, the model effectively performs question answering without prior context leading to enhanced zero-shot learning. Zero Redundancy Optimizer (ZeRO) achieves both model and data parallelism concurrently, which perhaps is the primary reason to train T-NLG with high throughput.

### V-F LANGUAGE MODELS ARE FEW-SHOT LEARNERS: GPT-III

The GPT family (I, II, and III) are autoregressive language models, based on transformer decoder blocks, unlike denoising autoencoder-based BERT. GPT-3 is trained on 175 billion parameters from a dataset of 300 billion tokens of text used for generating training examples for the model. Since GPT-3 is 10 times the size of any previous language model and for all tasks and purposes it employs few-shot learning via a text interface, without gradient updates or fine-tuning it achieves task agonism. It employs unsupervised pre-training, where the language model acquires a wide range of skills and pattern recognition capabilities. These are implemented on the fly to swiftly adapt to or identify the desired task. GPT-3 achieves SOTA in several NLP tasks although its few-shot learning falls short in reproducing similar results for other tasks.

### V-G SCALING GIANT MODELS WITH CONDITIONAL COMPUTATION AND AUTOMATIC SHARDING: GShard

GShard enables scaling beyond 600 billion parameters for multilingual machine translation via a sparsely gated mixture of experts (MoE) by automated sharding at low computation cost and compile time. The Transformer is sparsely scaled by inducting a position-wise mixture of experts (MoE) layer comprising of $E$ feed-forward networks $FFN_1, \ldots, FFN_E$ across its Transformer.

$$\mathcal{G}_{s,E} = GATE(x_s) \quad (32)$$
$$FFN_e(x_s) = wo_e.ReLU(wi_e.x_s) \quad (33)$$
$$y_s = \sum_{e=1}^{E} \mathcal{G}_{s,E}.FFN_e(x_s) \quad (34)$$

where $x_s$ and $y_s$ are the tokenized input and average weighted output to the MoE layer, $wi_e$ and $wo_e$ are an expert's (feed-forward layer) input and output projection matrices. The gating network indicates the expert's contribution to the final output via vector $\mathcal{G}_{s,E}$. This takes in a nonzero value for the tokens which are dispatched to a maximum of two experts that contribute to a non-zero value in an otherwise sparse matrix.

To achieve efficient parallelization across TPU clusters: (i) The parallelized attention layer is split along batch dimensions and weights are replicated across all devices. (ii) Due to size constraints, it's unfeasible to replicate MoE layer experts across all devices, hence experts are sharded across several devices, as shown below.

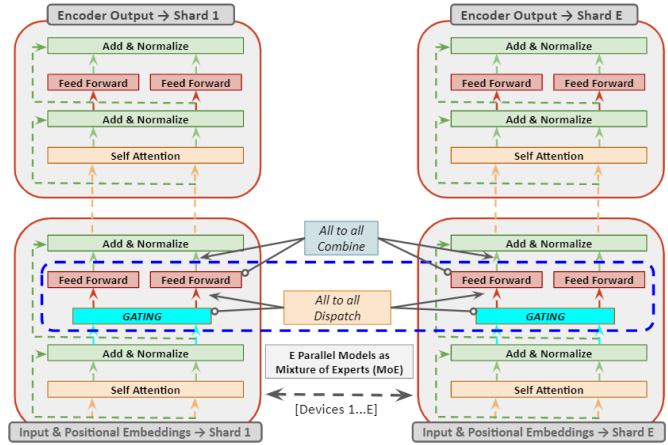

**FIGURE 14.** Sharded MoE Layered Transformer Encoder when scaled to multiple devices, all other layers are replicated

The two factors determining model quality are (i) High resourced languages where a vast amount of training data is available (ii) Enhancements for low-resourced languages with limited data. Increased tasks or language pairs in a translation model yields positive language transfer [69] for low-resource languages.

The three-pronged strategy for reasonable training time and efficiency for a large number of languages are: (i) Increase network depth by stacking more layers (ii) Increase network width by replicating the experts (iii) Sparsely assign tokens to experts via learned routing modules. When the number of experts per layer was quadrupled from 128 to 512 in a 12-layer deep model, a significant performance bump of 3.3 was observed in the BLEU score across 100 languages. Moreover, quadrupling the width from 512 to 2048 resulted in a diminishing gain in BLEU by 1.3. Further tripling the layer depth from 12 to 36 for the previously mentioned expert widths provides significant gains for low as well as high resources languages. However, increased model depth is not fruitful unless the model's capacity constraint (MoE width) is not relaxed.

## VI. MODEL SIZE REDUCTION
### VI-A DISTILLATION

The goal of Knowledge Distillation (KD) is to train a smaller student model under the supervision of a larger, more accurate teacher model via a revised loss function to achieve similar accuracy across unlabeled samples. The predicted teacher model samples are supplied to enable student learning through softer probabilistic class distribution while predicting through hard target classification via a separate loss function. This hard to soft label transition enables greater information variance for student learning, for instance, hard target classifies dog as $\{cow, dog, cat, car \in 0,1,0,0\}$ and soft target as $\{10^{-6}, 0.9, 0.1, 10^{-9}\}$. For hard classification computation, the last fully connected layer of a deep neural network is a vector of logits $z$, for which $z_i$ is the logit for the $i^{th}$ class. Therefore, probability $p_i$ that the input fits the $i^{th}$ class can be assessed by a softmax function in (35) and temperature element $T$ is inducted to influence each soft target's significance to be transferred to the student model learning in (36).

$$p_i = \frac{\exp(z_i)}{\sum_j \exp(z_j)} \quad (35); \quad p_i = \frac{\exp\left(\frac{z_i}{T}\right)}{\sum_j \exp\left(\frac{z_j}{T}\right)} \quad (36)$$

For a softer probability distribution over classes, a higher temperature is needed $(T = t)$. Experimentally it was discovered that it is fruitful to train the student model on correct (hard/ground truth) labels apart from teacher's soft labels. Although the student model cannot exactly match the soft targets, hard label training further assists it to not stumble to the incorrect prediction. The soft target distillation loss $(T = t)$ is computed by matching the logits between the teacher and the student model as:

$$\mathcal{L}_D\big(p(z_t, T), p(z_s, T)\big) = \sum_i -p_i(z_{ti}, T) \log\big(p_i(z_{si}, T)\big) \quad (37)$$

where $z_t$ and $z_s$ denote the logits of the teacher and student models, respectively. The distillation mechanism is clearly explained in figure 15. The cross-entropy between the ground truth label $y$ and the soft logits of the student model constitutes the student loss as:

$$\mathcal{L}_s(y, p(z_s, T)) = \sum_i -y_i \log(p_i(z_{si}, T)) \qquad (38)$$

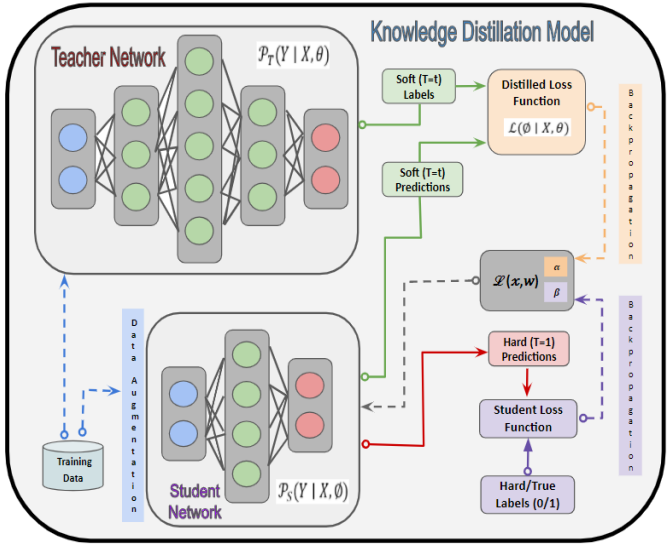

**FIGURE 15.** Language Model's Generalized Distilled Architecture

The standard model of vanilla knowledge distillation integrates the distilled and the student loss as shown below,
$\mathcal{L}(x, W) = \alpha \times \mathcal{L}_D(p(z_t, T), p(z_s, T)) + \beta \times \mathcal{L}_s(y, p(z_s, T))$ (39)
$(T = t) \text{ for } \mathcal{L}_D \text{ and } (T = 1) \text{ for } \mathcal{L}_s$
where $W \in$ student parameters and $\alpha, \beta \in$ regulated parameters. In the original paper weighted average was used concerning $\alpha$ and $\beta$, i.e., $\beta = 1 - \alpha$ and for best results, it was observed that $\alpha \gg \beta$.

### VI-A.1. DistilBERT
DistilBERT, the student version of the teacher BERT retained 97% of BERT's language understanding performance and was at inference time lighter, faster, and required lesser training cost. Through KD, DistilBERT reduces BERT size by 40%, is 60% faster and the compressed model is small enough to be operated on edge devices. The layer-depth of DistilBERT is slashed by half when compared with BERT since both possess the same dimensionality and possess generally an equivalent architecture. Layer reduction was performed as its normalization and linear optimization were computationally ineffective in the final layers. To maximize the inductive bias of large pre-trained models, DistilBERT introduced a triple loss function which linearly combined the distillation ($\mathcal{L}_D$) with the supervised training ($\mathcal{L}_{mlm}$) or the masked language modeling loss. It was observed that supplementing the prior loss with embedding cosine loss ($\mathcal{L}_{cos}$) was beneficial as it directionally aligned the teacher's and student's hidden state vectors.

### VI-A.2. TinyBERT
To overcome the distillation complexity of the pre-training-then-fine-tuning paradigm, TinyBERT introduced a lucid knowledge transfer process by inducting 3 loss functions: (i) Embedding Layer Output (ii) Attention Matrices, the Hidden States from Transformer (iii) Output Logits. This not only led TinyBERT to retain over 96% of BERT's performance at drastically reduced size but also deployed a meager 28% of parameters and 31% of inference time across all BERT-based distillation models. Further, it leveraged the untapped extractable potential from BERT's learned attention weights [70], for $(M + 1)^{th}$ layer, knowledge acquired is enhanced by minimizing:

$$\mathcal{L}_{model} = \sum_{m=0}^{M+1} \lambda_m \mathcal{L}_{layer}(S_m, T_{g(m)}) \qquad (40)$$

where $\mathcal{L}_{layer}$ is the loss function of a Transformer or an Embedding layer and hyperparameter $\lambda_m$ signifies the importance of $m^{th}$ layer's distillation. BERT's attention-based enhancement for language understanding can be incorporated in TinyBERT as:

$$\mathcal{L}_{attn} = \frac{1}{h} \sum_{i=1}^{h} MSE(A_i^S, A_i^T), \quad \text{where } A_i \in \mathbb{R}^{l \times l} \quad (41)$$

where $h$ denotes the number of heads, $A_i$ is the attention matrix corresponding to student or teacher's $i^{th}$ head, $l$ denotes input text length along with mean squared error (MSE) loss function. Further, TinyBERT distills knowledge from the Transformer output layer and can be expressed as:
$$\mathcal{L}_{hidn} = MSE(H^S W_h, H^T) \qquad (42)$$
where $W_h \in \mathbb{R}^{l \times d'}, H^S \in \mathbb{R}^{l \times d'}, H^T \in \mathbb{R}^{l \times d}, d' < d$
where $H^S, H^T$ are the hidden states of the student and teacher respectively, hidden sizes of the teacher and student models are denoted via scalar values of $d'$ and $d$, $W_h$ is a learnable matrix that transforms the student network's hidden states to the teacher network's space states. Similarly, TinyBERT also performs distillation on embedding-layer:
$$\mathcal{L}_{embd} = MSE(E^S W_e, E^T) \qquad (43)$$
where $E^S$ and $H^T$ are embedding matrices of student and teacher networks, respectively. Apart from mimicking the intermediate layer behavior, TinyBERT implements KD to fit predictions of the teacher model via cross-entropy loss between logits of the student and the teacher.

$$\mathcal{L}_{pred} = -softmax(z^T) \cdot \log\left(softmax\left(\left(\frac{z^S}{t}\right)\right)\right) \qquad (44)$$

Here $z^T$ and $z^S$ are the respective logits predicted by the teacher and student models.

### VI-A.3. MobileBERT
Unlike previous distilled models, MobileBERT achieves task-agnostic compression from BERT achieving training convergence via prediction and distillation loss. To train such a deeply thin model, a unique inverted bottleneck teacher model is designed that incorporates BERT (IB-BERT) from where knowledge transfer distills to MobileBERT. It is 4.3× smaller, 5.5× faster than BERT achieving a competitive score that is 0.6 units lower than BERT on GLUE-based inference tasks. Further, the low latency of 62 ms on Pixel 4 phone can be attributed to the replacement of Layer Normalization and gelu activation, with the simpler Hadamard product (∘) based linear transformation.

$$NoNorm(h) = Y \circ h + \beta, where\ Y, \beta \in \mathbb{R}^n \quad (45)$$

For knowledge transfer, the mean squared error between feature maps of MobileBERT's and IB-BERT is implemented as a transfer objective.

$$\mathcal{L}_{FMT}^l = \frac{1}{TN} \sum_{t=1}^{T} \sum_{n=1}^{N} (H_{t,l,n}^{tr} - H_{t,l,n}^{st})^2 \quad (46)$$

where $l$ is layer index, $T$ is sequence length, $N$ is the feature map size. For TinyBERT to harness the attention capability from BERT, KL-divergence is minimized between per-head distributions of the two models, where $A$ denotes the number of attention heads.

$$\mathcal{L}_{AT}^l = \frac{1}{TA} \sum_{t=1}^{T} \sum_{a=1}^{A} D_{KL}(a_{t,l,a}^{tr} || a_{t,l,a}^{st}) \quad (47)$$

Alternatively, a new KD loss can be implemented during MobileBERT's pre-training with a linear combination of BERT's MLM and NSP loss, where $\alpha$ is a hyperparameter between (0,1).

$$\mathcal{L}_{PD} = \alpha \mathcal{L}_{MLM} + (1-\alpha) \mathcal{L}_{KD} + \mathcal{L}_{NSP} \quad (48)$$

For the above-outlined objectives, 3 training strategies are proposed:

*(i) Auxiliary Knowledge Transfer:* Intermediary transfer via a linear combination of all layer transfer loss and distilled pre-training loss.

*(ii) Joint Knowledge Transfer:* For superior results, 2 separate losses are proposed where MobileBERT is trained with all layers that jointly transfer losses and perform pre-trained distillation.

*(iii) Progressive Knowledge Transfer:* To minimize error transfer from lower to higher layers, it is proposed to divide knowledge transfer into $L$ layered $L$ stages where each layer is trained progressively.

### VI-B PRUNING

Pruning [71] is a methodology where certain weights, biases, layers, and activations are zeroed out which are no longer a part of the model's backpropagation. This introduces sparsity in such elements which are visible post ReLU layer that converts negative values to zero $((ReLU(x): max(0,x))$. Iterative pruning learns the key weights, eliminating the least critical ones based on threshold values, and retraining the model enabling it to recuperate from pruning by adapting to the remaining weights. NLP models like BERT, RoBERTa, XLNet were pruned by 40% and retained their performance by 98%, which is comparable to DistilBERT.

#### VI-B.1 LAYER PRUNING
#### VI-B.1-A STRUCTURED DROPOUT

This architecture [72] randomly drops layers at training and test time that enables sub-network selection of any desired depth, since the network has been trained to be pruning robust. This is an upgrade from current techniques that require re-training a new model from scratch as opposed to training a network from which multiple shallow models are extracted. This sub-network sampling like Dropout [73] and DropConnect [74] builds an efficient pruning robust network if the smartly chosen simultaneous group of weights are dropped. Formally, pruning robustness in regularizing networks can be achieved by independently dropping each weight via Bernoulli's distribution where parameter p > 0 regulates the drop rate. This is comparable to the pointwise product of weight matrix $W$ with an arbitrarily sampled {0, 1) mask matrix $M$, $W_d = M \odot W$.

The most effective layer dropping strategy is to drop every other layer, where pruning rate $p$ and dropping layers at depth $d$ such that $d \equiv 0(mod \lfloor 1/p \rfloor)$. For $N$ groups with a fixed drop ratio $p$, the average number of groups utilized during training the network is $N(1-p)$, hence pruning size for $r$ groups, the ideal drop rate will be $p^* = 1 - r/N$. This approach has been highly effective on numerous NLP tasks and has led to models on size comparable to distilled versions of BERT and demonstrate better performance.

#### VI-B.1-B POOR MAN'S BERT

Due to the over-parameterization of deep neural networks, availability of all parameters is not required at inference time, hence few layers are strategically dropped resulting in competitive results for downstream tasks [75]. The odd-alternate dropping strategy drove superior results than the top and even alternate dropping for span $K = 2$ across all tasks. For instance, in a 12-layer network, dropping: top – {11, 12}; even-alternate – {10, 12}; odd-alternate – {9, 11}, concluded in (i) dropping the final two layers consecutively is more detrimental than eliminating alternate layers, and (ii) preserving the final layer has greater significance than other top layers.

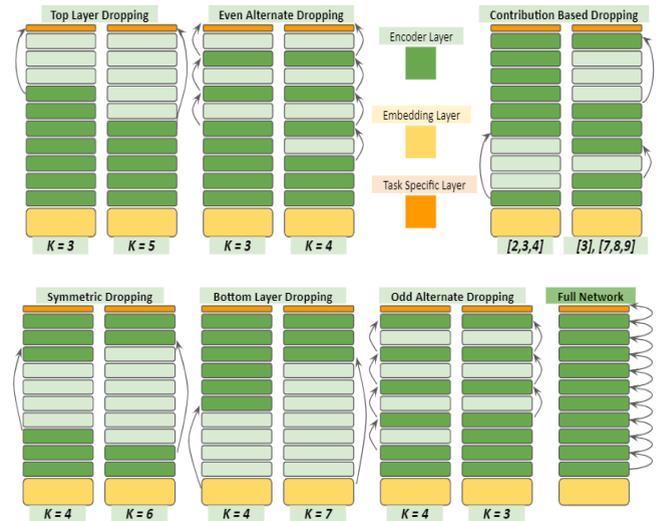

**FIGURE 16.** Layer Pruning Strategies deployed by Language Models

At higher values of $K$, the alternate dropping approach signifies a large drop in performance, hypothesized due to the elimination of lower layers. The Symmetric approach emphasizes the conservation of top and bottom layers while middle layers are dropped. This leads to a minimal impact on BERT while it substantially degrades XLNet's performance, resulting in the second-best strategy for BERT giving robust results even after removal of 4 layers.

Observationally XLNet demonstrates greater pruned robustness than BERT as its learning mellows close to its 7th layer whereas BERT keeps learning until the 11th layer. Consequently (i) XLNet gathers task-oriented knowledge at lower layers in contrast to BERT, (ii) XLNet's final layers might get fairly redundant and are liable to get dropped without a considerable drop in performances. The authors furthered the dropping experimentations to DistilBERT, here dropping 30% of its layers resulted in minimal performance degradation.

Like previous models, top-layer dropping turned out to be most reliable as RoBERTa proved to be more pruning robust than BERT as a 6-layered RoBERTa demonstrated similar performances to DistilRoBERTa. All the layer dropping strategies can be visualized from the above figure 16.

### VI-B.2 WEIGHT PRUNING

Prior work focuses primarily on unstructured individual weight pruning [76], [77], although effective its resulting unstructured sparse matrices are challenging to process on conventional hardware. This makes it difficult to secure inference speedups despite model size reduction. Contrarily structured pruning enforces highly structured weight matrices which when optimized via dense linear algebraic implementation, lead to substantial speedup but lower performance than unstructured pruning due to greater constraints.

### VI-B.2-A STRUCTURED PRUNING

To overcome the above shortcomings, a novel structured pruning paradigm was introduced [78] with low-rank factorization which retained the dense matrix structure and $l_0$ norm which relaxed constraints enforced via structured pruning. The weight matrices were factorized into a product of two smaller matrices with a diagonal mask that was pruned while training via $l_0$ regularizer that controlled the end sparsity of the model. This generic method FLOP (Factorized $L0$ Pruning) could be employed for any matrix multiplication. For a neural network $f(;\boldsymbol{\theta})$ parameterized by $\boldsymbol{\theta} = \{\theta_j\}_{j=1}^n$ where each $\theta_j$ represents an individual weight or a block of weights (e.g., column matrix) and $n$ denotes the number of blocks. Consider a pruning binarized variable $z = \{z_j\}_{j=1}^n$ where $z_j \in \{0,1\}$, $\tilde{\boldsymbol{\theta}} = \{\tilde{\theta}_j\}$ denotes model parameter set, post pruning via $l_0$ normalization.

$$\tilde{\theta} = \theta \odot z \quad \forall j \ \tilde{\theta}_j = \theta_j z_j \quad (49)$$

Consider a matrix $W$ to be factorized into a product of two smaller matrices $P$ and $Q$ where $W = P.Q$ and $r$ is the number of $P$ columns or $Q$ rows. Structured Pruning for each component is attained via a pruning variable $z_k$

$$W = PGQ = \sum_{k=1}^{r} z_k \times (p_k \times q_k) \quad (50)$$

where $G = diag(z_1, ..., z_r)$

### VI-B.3 HEAD PRUNING

Though certain models have a greater dependency on multiple heads in a multi-headed attention environment, recent work reveals that a significant portion of attention heads can be removed resulting in a pruned model with enhanced memory efficiency, speed, and accuracy. Prior work [79],[80] judged head importance via averaging the attention weights over all heads at a particular position or based their results on maximum attention weight values. However, both approaches did not unequivocally consider the fluctuating significance of different heads.

### VI-B.3-A Analyzing Multi-Head Self-Attention: Specialized Heads Do the Heavy Lifting, Rest Can Be Pruned

This model [81] unearthed three distinct layered head roles: *(i)* <u>Positional heads</u>: Attending to an adjacent token *(ii)* <u>Syntactic heads</u>: Attending to those with syntactic dependency *(iii)* <u>Rare Word heads</u>: Indicating to least frequent tokens in a sequence. Based on the above roles [81] the revelations are summarized as *(a)* Small subset of heads were key for translation *(b)* Key heads possessed a single, often more specialized and interpretable model functionality *(c)* The head roles corresponded to adjacent tokens attention in an explicit syntactic dependency relation. High head confidence via Layer-wise Relevance Propagation (LRP) [82] relates to the proportion of a token's attention defined as the median of its maximum attention weight computed over all tokens, which is expected to be crucial for a task. The modified Transformer architecture via product of each head's computed representation $head_i$ and scalar gate $g_i$, $MultiHead(Q,K,V) = Concat_i(g_i.head_i)W^O$, where $g_i$ are input independent head specific parameters, $L_0$ regularization is applied to $g_i$ for less important heads that need to be disabled, where $h \in (number\ of\ heads)$.

$$L_0(g_1 ... g_h) = \sum_{i=1}^{h}(1 - [[g_i = 0]]) \quad (51)$$

However, $L_0$ norm is non-differentiable; hence it cannot be inducted as a regularization term in the objective function. Therefore, a stochastic relaxation is applied where each gate $g_i$ is randomly picked from a head distribution obtained via stretching (0,1) to $(-\epsilon, 1+\epsilon)$ and collapsing the probability distribution $(-\epsilon, 1]$ to $[1, 1+\epsilon)$ to singular points 0 and 1. This rectified stretching results in a distribution over [0,1] that is mixed discretized-continuous. The probability sum of heads being non-zero can be implemented as a relaxed L0 norm.

$$\mathcal{L}_C(\emptyset) = \sum_{i=1}^{h}(1 - P(g_i = 0 \mid \emptyset_i)) \quad (52)$$

The modified training regime can be expressed as $\mathcal{L}(\theta, \emptyset) = \mathcal{L}_{xent}(\theta, \emptyset) + \lambda \mathcal{L}_C(\emptyset)$, where $\theta$ are original Transformer's parameters, $\mathcal{L}_{xent}(\theta, \emptyset)$ is the translation model's cross-entropy loss and $\mathcal{L}_C(\emptyset)$ is the regularizer.

### VI-B.3-B ARE 16 HEADS REALLY BETTER THAN ONE?

In multi-headed attention (MHA), consider a sequence of $n$ $d$-dimensional vectors $x = x_1,..,x_n \in \mathbb{R}^d$, and query vector $q \in \mathbb{R}^d$. The MHA layer parameters $W_q^h, W_k^h W_v^h W_o^h \in \mathbb{R}^{d_h \times d}$ and $W_o^h \in \mathbb{R}^{d \times d_h}$, when $d_h = d$. For masking attention heads the original transformer equation is modified as:

$$MH_{Attn}(x,q) = \sum_{h=1}^{N_h} \xi_h Att_{W_q^h, W_k^h W_v^h W_o^h}(x,q) \quad (53)$$

where $\xi_h$ are masking variables with values between $\{0,1\}$, $Att_h(x)$ is the output of head $h$ for input $x$. The following experiments yielded the best results [83] on pruning the different number of heads at test times:

*(i)* *Pruning just one head:* If the model's performance significantly degrades while masking head $h$, then $h$ is a key head else it is redundant given the rest of the model. A mere 8 (out of 96) heads trigger a significant change in performance when removed from the model, out of which half result in a higher BLEU score.

*(ii)* *Pruning all heads except one:* A single head for most layers was deemed sufficient at test time, even for networks with 12 or 16 attention heads, resulting in a drastic parametric reduction. However, multiple attention heads are a requirement for specific layers i.e., the final layer of the encoder-decoder attention, where performance degrades by a massive 13.5 BLEU points on a single head.

The expected sensitivity of the model to the masking $\xi$ is evaluated for the proxy score for head significance.

$$I_h = \mathbb{E}_{x \sim X} \left| \frac{\partial \mathcal{L}(x)}{\partial \xi_h} \right| \quad (54)$$

$$I_h = \mathbb{E}_{x \sim X} \left| Att_h(x)^T \frac{\partial \mathcal{L}(x)}{\partial Att_h(x)} \right| \quad (55)$$

where $X$ is the data distribution, $\mathcal{L}(x)$ is the loss on sample $x$. If $I_h$ is high, then modifying $\xi_h$ will likely have a significant effect on the model, hence low $I_h$ value heads are iteratively pruned out.

### VI-C QUANTIZATION

32-bit floating-point (FP32) has been the predominant numerical format for deep learning, however the current surge for reduced bandwidth and compute resources has propelled the implementation of lower-precision formats. It has been demonstrated that weights and activation representations via 8-bit integers (INT8) have not led to an evident accuracy loss. For instance, BERT's quantization to 16/8-bit weight format resulted in 4× model compression with minimal accuracy loss, consequently, a scaled-up BERT serves a billion CPU requests daily.

### VI-C.1 LQ-NETS

This model [84] inducts simple to train network weights and activations mechanism via joint training of a deep neural network. It quantizes with variable bit precision capabilities unlike fixed or manual schemes [85],[86]. Generally, a quantized function can represent floating-point weights $w$, activations $a$, in a few bits as:

$$Q(x) = q_l, \text{ if } x \in (t_l, t_{l+1}] \text{ where } q_l, l = (1,...,L) \quad (56)$$

Here $q_l$ and $(t_l, t_{l+1}]$ are quantization levels and intervals, respectively. To preserve quick inference times, quantization functions need to be compatible with bitwise operations, which is achieved via uniform distribution that maps floating-point numbers to their nearest fixed-point integers with a normalization factor. The LQ learnable quantization function can be expressed as:

$$Q_{LQ}(x,v) = v^T e_l, \text{ if } x \in (t_l, t_{l+1}] \quad (57)$$

where $v \in \mathbb{R}^K$ is the learnable floating-point basis and $e_l \in \{-1,1\}^K$ for $l = (1,..,L)$ enumerating $K$-bit binary encodings from $[-1,..,-1]$ to $[1,..,1]$. The inner product computation of quantized weights and activations is computed by the following bitwise operations with weight bit-width $K_w$.

$$Q_{LQ}(w, v^w)^T Q_{LQ}(a, v^a) = \sum_{i=1}^{K_w} \sum_{j=1}^{K_a} v_i^w v_j^a (b_i^w \odot b_j^a) \quad (58)$$

where $w, a \in \mathbb{R}^n$ encoded by vectors $b_i^w, b_j^a \in \{-1,1\}^N$ where $i = 1,...,K_w$ and $j = 1,...,K_a$ and $v^w \in \mathbb{R}^{K_w}, v^a \in \mathbb{R}^{K_a}$, $\odot$ denotes bitwise inner product $xnor$ operation.

### VI-C.2 QBERT

QBERT [87] deploys a two-way BERT quantization with input $x \in X$, its corresponding label y $\in Y$, via cross entropy-based loss function

$$L(\theta) = \sum_{(x_i, y_i)} CE(softmax(W_c(W_n(...W_1(W_e(x_i))))), y_i) \quad (59)$$

where $W_e$ is the embedding table, with encoder layers $W_1, W_2...W_n$ and classifier $W_c$. Assigning the same bit size representation to different encoder layers with varying sensitivity attending to different structures [5] is sub-optimal and it gets intricate for small target size (2/4 bits) requiring ultra-low precision. Hence via Hessian Aware Quantization (HAWQ) more bits are assigned to greater sensitive layers to retain performance. Hessian matrix is computed via computationally economical matrix-free iteration technique where first layer encoder gradient $g_1$ for an arbitrary vector $v$ as:

$$\frac{\partial g_1^T v}{\partial W_1} = \frac{\partial g_1^T}{\partial W_1} v + g_1^T \frac{\partial v}{\partial W_1} = \frac{\partial g_1^T}{\partial W_1} v = H_1 v \quad (60)$$

where $H_1$ is Hessian matrix of the first encoder and $v$ is independent to $W_1$, this approach determines the top eigenvalues for different layers and more aggressive quantization is deployed for layers with smaller eigenvalues. For further optimization via group-wise quantization, each dense matrix is treated as a group with its quantization range and is partitioned following each continuous output neuron.

### VI-C.3 Q8BERT

To quantize weights and activations to 8-bits, symmetric linear quantization is implemented [88], where $S^x$ is the quantized scaling factor for input $x$ and ($M = 2^{b-1} - 1$) is the highest quantized value when quantizing to $b$ bits.
$$Quantize(x|S^x, M) := Clamp(\lfloor x \times S^x \rfloor, -M, M) \quad (61)$$
$$Clamp(x, a, b) = \min(\max(x, a), b)$$
Implementing a combination of fake quantization [89] and Straight-Through Estimator (STE) [90], inference time quantization is achieved during training with a full-precision backpropagation enabling FP32 weights to overcome errors. Here $\frac{\partial x^q}{\partial x} = \vec{1}$, where $x^q$ is the result of fake quantizing $x$.

## VII. INFORMATION RETRIEVAL

For knowledge-intensive tasks like efficient data updating, and retrieval, huge implicit knowledge storage is required. Standard language models are not adept at these tasks and do not match up with task-specific architectures which can be crucial for open-domain Q&A. For instance, BERT can predict the missing word in the sentence, "The ___ is the currency of the US" (answer: "dollar"). However since this knowledge is stored implicitly in its parameters, the size substantially increases to store further data. This constraint raises the network latency and turns out prohibitively expensive to store information as storage space is limited due to the size constraints of the network.

### VII-A GOLDEN RETRIEVER

A conventional multi-hop based open-domain QA involves question $q$ and from a large corpus containing relevant contextual $S$ (gold) documents $d_1, \ldots, d_s$ that form a sequence of reasoning via textual similarities that lead to a preferred answer $a$. However, GoldEn Retriever's [91] first-hop generates a search query $q_1$ that retrieves document $d$ for a given question $q$, thereafter for consequent reasoning steps ($k = 2, \ldots, S$) a query $q_k$ is generated from the question ($q$) and available context ($d_1, \ldots, d_{k-1}$). GoldEn retrieves greater contextual documents iteratively while concatenating the retrieved context for its QA model to answer. It is independent of the dataset and task-specific IR models where indexing of additional documents or question types leads to inefficiencies. A lightweight RNN model is adapted where text spans are extracted from contextual data to potentially reduce the large query space. The goal is to generate a search query $q_k$ that helps retrieve $d_k$ for the following reasoning step, based on a textual span from the context $C_k$, $q$ is selected from a trained document reader.
$$q_k = G_k(q, C_k) \quad (62), \quad C_{k+1} = C_k \text{ concat } IR_n(q_k) \quad (63)$$
where $G_k$ is the query generator and $IR_n(q_k)$ are top n retrieved documents via $q_k$.

### VII-B ORQA

The components reader and retriever are trained jointly in an end-to-end fashion where BERT is implemented for parameter scoring. It can retrieve any text from an open corpus and is not constrained by returning a fixed set of documents like a typical IR model. The retrieval score computation is the question's $q$ dense inner product with evidence block $b$.
$$h_q = W_q BERT_Q(q)[CLS] \quad (64)$$

$h_b = W_b BERT_B(b)[CLS]$ (65), $S_{retr}(b,q) = h_q^T h_b$ (66)
where $W_q$ and $W_b$ matrices project the BERT output into 128-dimensional vectors. Similarly, the reader is BERT's span variant of the reading model.
$$h_{start} = BERT_R(q,b)[START(s)], \quad (67)$$
$$h_{end} = BERT_R(q,b)[END(s)], \quad (68)$$
$$S_{read}(b,s,q)MLP([h_{start}; h_{end}]) \quad (69)$$
The retrieval model is pre-trained with an Inverse Cloze Task (ICT), where the sentence context is relevant semantically and is used to extrapolate data missing from the sequence $q$.
$$P_{ICT}(b|q) = \frac{\exp(S_{retr}(b,q))}{\sum_{b' \in BATCH} \exp(S_{retr}(b',q))} \quad (70)$$
where $q$ is treated as pseudo-question, $b$ is text encircling $q$ and $BATCH$ is a set of evidence blocks employed for sampling negatives. Apart from learning word matching features, it also learns abstract representations as pseudo-question might or might not be present in the evidence. Post ICT, learning is defined distribution over answer derivations.
$$P_{learn}(b,s|q) = \frac{\exp(S(b,s,q))}{\sum_{b' \in TOP(k)} \sum_{s' \in b'} \exp(S(b',s',q))} \quad (71)$$
where $TOP(k)$ are top retrieved blocks based on $S_{retr}$. In this framework, evidence retrieval from complete Wikipedia is implemented as a latent variable which is unfeasible to train from scratch hence retriever is pre-trained with an ICT.

### VII-C REALM

This framework explicitly attends to a vast corpus like Wikipedia however, its retriever learns via backpropagation and performs Maximum Inner Product Search (MIPS) via cosine similarity to chose document appropriateness. The retriever is designed to cache and asynchronously update each document to overcome the computational challenge of multi-million order retrieval of candidate documents.
In pre-training, the model needs to predict the randomly masked tokens via the knowledge retrieval relevance score $f(x,z)$, the inner product of vector embeddings between $x$ and $z$ (MIPS). To implement a knowledge-based encoder, the combination of input $x$ and retrieved document $z$ from a corpus $\mathbb{Z}$ is fed as a sequence to fine-tune the Transformer $p(y | z, x)$ as shown in figure 17. This enables complete cross attention between $x$ and $z$ that enables to predict the output $y$ where:
$$f(x,z) = Embed_{Input}(x)^T Embed_{doc}(z)$$
$$p(z | x) = \frac{\exp f(x,z)}{\sum_{z'} \exp f(x,z')} \quad (72)$$
$$p(y | x) = \sum_{z \in \mathbb{Z}} p(y | z, x) p(z | x) \quad (73)$$
Like ORQA, BERT is implemented for embedding:
$$join_{BERT}(x) = [CLS]x[SEP] \quad (74)$$
$$join_{BERT}(x_1, x_2) = [CLS]x_1[SEP]x_2[SEP] \quad (75)$$
In the pre-training of the BERT's masked language modeling task, each mask in token $x$ needs to be predicted as:

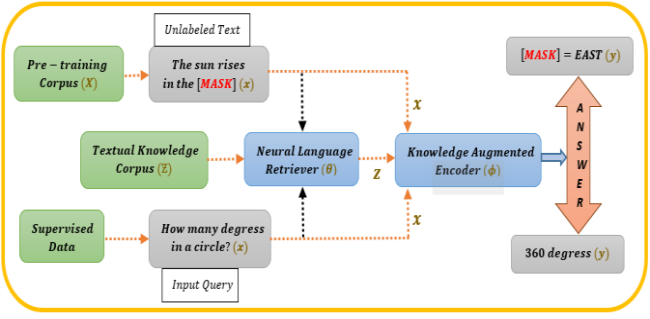

**FIGURE 17.** Unsupervised pre-training (top) and Supervised fine-tuning (bottom) in REALM's architecture

$$p(y \mid z, x) = \prod_{j=1}^{J_x} p(y_j \mid z, x) \quad (76)$$

$$p(y_j \mid z, x) \propto$$
$$\exp\left(w_j^T BERT_{MASK(j)}\left(join_{BERT}(x, z_{body})\right)\right) \quad (77)$$

where $BERT_{MASK(j)}$ represents the Transformer output vector corresponding to the $j^{th}$ masked token. $J_x$ is the total number of $[MASK]$ tokens in $x$, and $w_j$ is the learned word embedding for token $y_j$. For an open-ended Q&A fine-tuning task, answer $y$ is in the form of a spanned token sequence in the target document $z$. The span set $S(z, y)$ matching $y$ in $z$ can be modeled as:

$$p(y \mid z, x) \propto \sum_{s \in S(z,y)} \exp\left(MLP([h_{START(s)}; h_{END(s)}])\right) \quad (78)$$

$$h_{START(s)} = BERT_{START(s)}\left(join_{BERT}(x, z_{body})\right) \quad (79)$$

$$h_{END(s)} = BERT_{END(s)}\left(join_{BERT}(x, z_{body})\right) \quad (80)$$

where $BERT_{START(s)}$ and $BERT_{END(s)}$ denote the Transformer output vectors corresponding to the start and end tokens of span $S$ and $MLP$ denotes a feed-forward neural network.

### VII-D RETRIEVAL AUGMENTED GENERATION: RAG

RAG is a flexible combination of the 'closed-book' i.e., parametric model and the performance of 'open-book' i.e., retrieval model approaches, outperforming current language models. A parametric memory is a sequence to sequence pre-trained model whereas a Wikipedia representation via a dense vector index constitutes non-parametric memory, which is accessed via a pre-trained neural retriever. Since RAG is built as a culmination of the two it does not require prior training since knowledge is available via retrieved pre-trained data unlike former non-parametric architectures [92]. To achieve greater context in output sequence $(y)$ generation, the general-purpose RAG incorporates retrieved text passages $z$ for a given input $x$, that involve two major components:

(i) Retriever $p_\eta(z \mid x)$, parameterized via $\eta$, it returns the top matched content from text passages for query $x$, this $RAG\ Sequence$ architecture's retrieved passage acts as a latent variable marginalized to achieve maximum probability $p(y \mid x)$ across top-K approximations.

$$p_{RAG-Sequence}(y \mid x) = \sum_{z \in top-k(p(.\mid X))} p_\eta(z \mid x) \prod_i^N p_\theta(y_i \mid x, z, y_{1:i-1}) \quad (81)$$

(ii) Generator $p_\theta(y_i \mid x, z, y_{1:i-1})$, parameterized via $\theta$, it generates the current token $y_i$ based on a contextual representation of the prior $i - 1$ tokens $y_{1:i-1}$, input $x$ and retrieved passage $z$. The $RAG\ Token$ model predicts each target token based on a different latent passage, simultaneously enabling the generator to select subject matter from various documents.

$$p_{RAG-Token}(y \mid x) = \prod_i^N \sum_{z \in top-k(p(.\mid X))} p_\eta(z_i \mid x) p_\theta(y_i \mid x, z_i, y_{1:i-1}) \quad (82)$$

The retrieval module $p_\eta(z \mid x)$ is based on Dense Passage Retrieval (DPR) where $d(z)$ the document's dense representation generated via BERT and $q(x)$ the query representation generated via another BERT.

$$p_\eta(z \mid x) \propto exp\langle d(z), q(x) \rangle \quad (83)$$

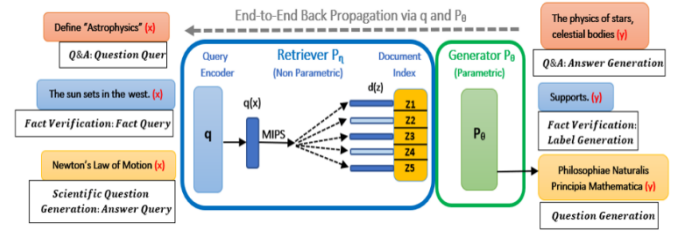

**FIGURE 18.** RAG's Parametric and Retrieval Model architecture

To effectively compute $top - k(p_\eta(.\mid x))$ elements $z$ with the highest probability $p_\eta(z \mid x)$ DPR employs a MIPS index where BART is used as the generator $p_\theta(y_i \mid x, z_i, y_{1:i-1})$. The retriever and generator are trained in conjunction to retrieve the target document in a semi-unsupervised manner.

### VII-D DENSE PASSAGE RETRIEVAL: DPR

DPR enhances open-domain QA retrieval using the dual encoder approach, unlike the computationally intensive ICT. Its dense encoder $E_P(\cdot)$ indexes all $M$ passages in a continuous, low-dimensional $(d)$ space that could effectively retrieve top relevant passages for a query at run time. A separate encoder $E_Q(\cdot)$ is deployed for the query and d-dimensional vector to map at run time, that retrieves $k$ passages which are most relevant to the question vector. The dot product computation between the query and passage determines their similarity. $sim(q, p) = E_Q(q)^T . E_Q(q)$. The goal is to learn a superior embedding function via training encoders that involve the creation of vector space where the relevant question, passage pairs possess smaller distances i.e., greater similarity than irrelevant ones. Assume training data with $m$ instances $\mathcal{D} = \{\langle y_i, p_i^+, p_{i,1}^-, --, p_{i,n}^- \rangle\}_{i=1}^m$ where each instance contains one query $q_i$, one positive (relevant) passage $p_i^+$ with $n$ negative (irrelevant) passages $p_{i,j}^-$. The loss function can be optimized as the negative log-likelihood of the positive passage.

$L(q_i, p_i^+, p_{i,1}^-, ..., p_{i,n}^-) =$

$$-log \frac{e^{sim(q_i, p_i^+)}}{e^{sim(q_i, p_i^+)} + \sum_{j=1}^{n} e^{sim(q_i, p_{i,j}^-)}} \quad (84)$$

## VIII. LONG SEQUENCE MODELS

Vanilla Transformers break input sequences into chunks if their length exceeds 512 tokens, which results in loss of context when related words exist in different chunks. This constraint results in a lack of contextual information leading to inefficient prediction and compromised performance and dawned the rise of such models.

### VIII-A DEEPER SELF-ATTENTION

This 64 layered Transformer [93] was built based on the discovery that it possessed greater character level modeling of longer-range sequences. The information was swiftly transmitted over random distances as compared to RNN's unitary step progression. However, the three following supporting loss parameters were added to the vanilla Transformer which accelerated convergence and provided the ability to train deeper networks.

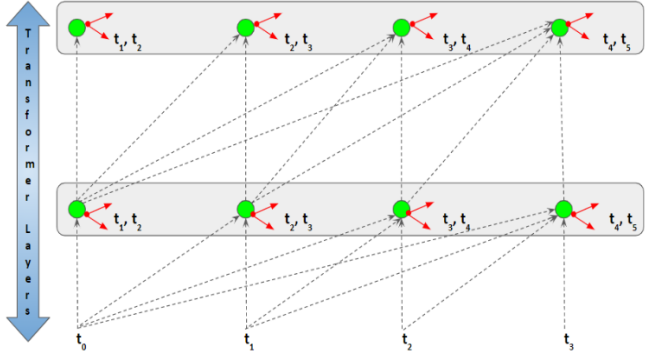

**FIGURE 19.** Accelerated convergence via multiple target token prediction across multiple positions through intermediate layers

(i) *Prediction across Multiple positions*: Generally causal prediction occurs at a single position in the final layer, however in this case all positions are used for prediction. These auxiliary losses compel the model to predict on smaller contexts and accelerate training without weight decay.
(ii) *Predictions on Intermediate Layer*: Apart from the final layer, predictions from all intermediate layers are added for a given sequence, as training progresses, lower layers weightage is progressively reduced. For $n$ layers, the contribution of $l^{th}$ intermediate layer ceases to exist after completing $l/2n$ of the training.
(iii) *Multiple Target Prediction*: The model is modified to generate two or greater predictions of future characters where a separate classifier is introduced for every new target. The extra target losses are weighed in half before being added to a corresponding layer loss.

The above 3 implementations are expressed in figure 19. For sequence length $L$, the language model computes joint probability autoregressive distribution over token sequences.

$$P(t_{0:L}) = P(t_0) \prod_{i=1}^{L} P(t_i \mid t_{0:i-1}) \quad (85)$$

### VIII-B TRANSFORMER-XL

To mitigate *context fragmentation* in vanilla Transformers, XL incorporates lengthier dependencies where it reuses and caches the prior hidden states from where data is propagated via recurrence. Given a corpus of tokens $x = (x_1, x_2 ..., x_T)$, a language model computes the joint probability $P(x)$ autoregressively, where the context $x_{<t}$ is encoded into a fixed size hidden state.

$$P(x) = \prod_{t} P(x_t \mid x_{<t}) \quad (86)$$

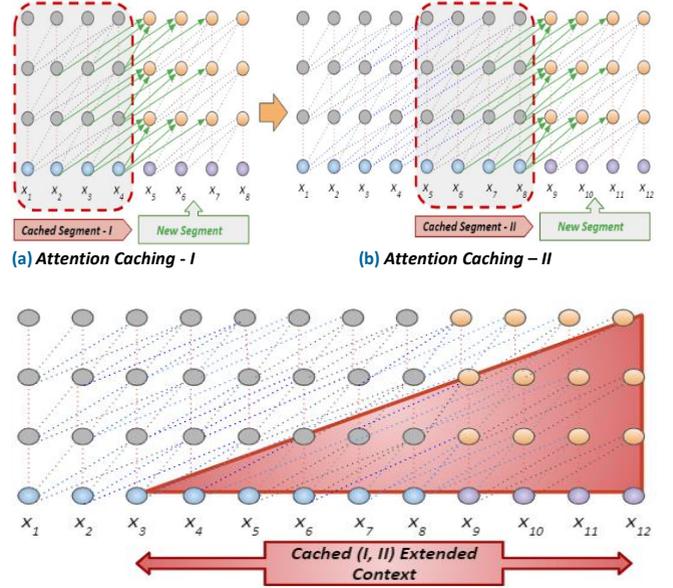

**FIGURE 20.** Elongated context capture combining (a) and (b)

Assume two consecutive sentences of length $L$, $s_\tau = [x_{\tau,1}, ..., x_{\tau,L}]$ and $s_{\tau+1} = [x_{\tau+1,1}, ..., x_{\tau+1,L}]$ where $n^{th}$ layer hidden state sequence produced by the $\tau^{th}$ segment $s_\tau$ as $h_\tau^n \in R^{L \times d}$, where $d$ is the hidden dimension. The $n^{th}$ hidden layer state for the segment $s_{\tau+1}$ is computed as follows:

$$\tilde{h}_{\tau+1}^{n-1} = [SG(h_\tau^{n-1}) \bullet h_{\tau+1}^{n-1}] \quad (87)$$
$$q_{\tau+1}^n, k_{\tau+1}^n, v_{\tau+1}^n = h_{\tau+1}^{n-1} W_Q^T, \tilde{h}_{\tau+1}^{n-1} W_K^T, \tilde{h}_{\tau+1}^{n-1} W_V^T \quad (88)$$
$$h_{\tau+1}^n = Transformer - Layer(q_{\tau+1}^n, k_{\tau+1}^n, v_{\tau+1}^n) \quad (89)$$

where $SG(\cdot)$ represents *stop-gradient*, $[h_u \bullet h_v]$ is the two hidden sequence concatenation, and $W$ the model parameters. The key distinction from the original Transformer lies in modeling the key $k_{\tau+1}^n$ and value $v_{\tau+1}^n$ concerning the extended context $h_{r+1}^{\sim n-1}$ and hence preceding $h_r^{n-1}$ are cached. This can be demonstrated from figure 20 above where prior attention span is cached by the latter forming an elongated caching mechanism.

Such recurrence is applied to every two consecutive segments to create a segment level recurrence via hidden states. In the original transformer the attention score within the same segment between query ($q_i$) and key ($k_i$) vector is:
$$A_{i,j}^{abs} = E_{x_i}^T W_q^T W_k E x_j + E_{x_i}^T W_q^T W_k U_j + \quad (90)$$
$$U_i^T W_q^T W_k E x_j + U_i^T W_q^T W_k U_j$$
From a perspective of relative positional encoding, the above equation is remodeled in the following manner
$$A_{i,j}^{rel} = E_{x_i}^T W_q^T W_{k,E} E x_j + E_{x_i}^T W_q^T W_{k,R} \mathbf{R_{i-j}} + \quad (91)$$
$$\mathbf{u}^T W_{k,E} E x_j + \mathbf{v}^T W_{k,R} \mathbf{R_{i-j}}$$

### VIII-C LONGFORMER
This architecture provides sparsity to the full attention matrix while identifying input location pairs attending one another and implements three attention configurations:
(i) <u>Sliding Window</u>: For a fixed window size $w$, each token attends to a sequence length (n) of $w/2$ on either side. This leads to the computational complexity of $O(n \times w)$ that scales linearly with input sequence length and for efficiency purposes $w < n$. A stacked $'l'$ layered transformer enables receptivity sized $'l \times w'$ over the entire input $'w'$ across all layers. Different $'w'$ values can be chosen for efficiency or performance.
(ii) <u>Dilated Sliding Window</u>: To conserve computation and extend the receptive field size to $'l \times d \times w'$, where $'d'$ variable-sized gaps are inducted for dilations in window size $'w'$. Enhanced performance is achieved via enabling few dilation-free heads (smaller window size) for attention on local context (lower layers) and remaining dilated heads (increased window size) attending longer context (higher layers).
(iii) <u>Global Attention</u>: The prior two implementations do not possess enough flexibility for task-precise learning. Hence "$global\ attention$" is implemented on few pre-designated input tokens $(n)$ where a token attends to all sequence tokens and all such tokens attend to it. This preserves the local and global attention complexity to $O(n)$.

Its attention complexity is the sum of local and global attention versus RoBERTa's quadratic complexity which is explained by the following mathematical expressions.

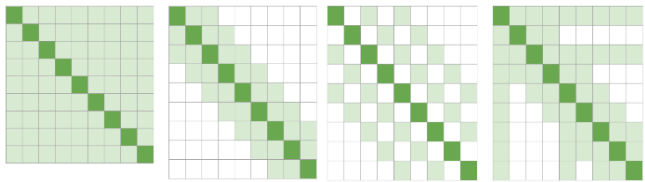

(a) **Full Attention**  (b) **Sliding Window**  (c) **Dilated Sliding**  (d) **Global Attention**

**FIGURE 21.** Longformer's different Sparse Attention configurations

$local\ attention = (n \times w)$

$where\ n \in (input\ sequence\ size), w \in (window\ size)$
$global\ attention = (2 \times n \times s)$
$where\ s \in (number\ of\ tokens\ with\ full\ attention)$
$Window\ Attention\ Size = n_0, hence\ (n_0 = w)$
$Total\ attention\ complexity = n(n_0 + 2s)\ \varepsilon\ O(n)$
$if\ n_0 \neq n$
$Total\ Memory\ Requirements =$
$\quad n(n_0 + 2s) \times Number\ of\ Transforer\ Layers$

Global attention enables chunk-less document processing, however, its space-time complexity will be greater than RoBERTa, if sequence length exceeds the window size.
$$\{O(RoBERTa) = O(n_0)^2\} < \begin{Bmatrix} O(Longformer) \\ = O(n(n_0 + 2s)) \end{Bmatrix}$$
$if\ n > n_0$

### VIII-D EXTENDED TRANSFORMER CONSTRUCTION: ETC
ETC is an adaptation of the Longformer design which receives global ($n_g$) and long ($n_l$) inputs where $n_g \ll n_l$. It computes four global-local attention variations: global-to-global ($g2g$), global-to-long ($g2l$), long-to-global ($l2g$), and long-to-long ($l2l$) to achieve long sequence processing. Global inputs and the other three variations possess limitless attention to compensate for $l2l's$ fixed radius span to achieve a balance between performance and computational cost. Further, it replaces the absolute with relative position encodings which provide information of input tokens concerning each other.

### VIII-E BIG BIRD
Mathematically Big Bird proves randomly sparse attention can be Turing complete and behaves like a Longformer aided with random attention. It is designed such as *(i)* a global token group $g$ attending to all sequence parts *(ii)* there exists a group of $r$ random keys that each query $q_i$ attends to *(iii)* a local neighbor window $w$ block that each local node attends to. Big Bird's global tokens are constructed using a two-fold approach *(i) Big Bird-ITC*: Implementing Internal Transformer Construction (ITC) where few current tokens are made global that attend over the complete sequence. *(ii) Big Bird-ETC*: Implementing Extended Transformer Construction (ETC), essential additional global tokens $g$ are included [$CLS$] that attend to all existing tokens.

Its definitive attention process consists of the following properties: queries attend to $r$ random keys where each query attends to $w/2$ tokens to the left and right of its location and have $g$ global tokens which are derived from current tokens or can be supplemented when needed.

## IX. COMPUTATIONALLY EFFICIENT ARCHITECTURES
### IX-A SPARSE TRANSFORMER
This model's economical performance is due to the alienation from the full self-attention procedure that is modified across several attention steps. The model's output results are derived from a factor of the full input array i.e., ($\sqrt{N}$) where $N \dot\in Sequence\ Length$ as expressed in Figure 22. This leads to a lower attention complexity of $O(N\sqrt{N})$ in contrast to Transformer's $O(N^2)$. Further, it deciphers sequences thirty times longer than its predecessors. Its factorized self-attention consists of $p$ distinct heads where the $m^{th}$ head defines a subset of attention indices $A_i^{(m)} \subset \{j : j \leq i\}$ and to generate sparsity $|A_i^{(m)}| \propto \sqrt[p]{n}$ leads to efficient choices for set $A$.

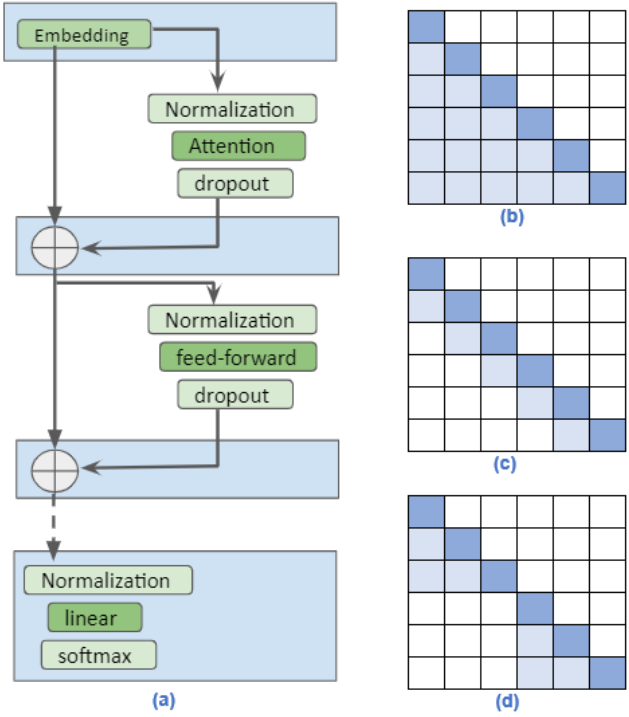

**FIGURE 22. (a)** Sparse Transformer Architecture **(b)** Decoder based Full Attention with Causal Masking **(c)** Stridden Sparsity **(d)** Fixed Sparsity

The *strided* attention is implemented in two dimensions where one head attends to previous $l$ locations and the other attends to each $l^{th}$ location, where stride $l$ value is close to $\sqrt{n}$. This is expressed as $A_i^{(1)} = \{t, t+1, ..., i\}$ for $t = max(0, i-l)$ and $A_i^{(1)} = \{j : (i-j) \mod l = 0\}$. This linear transformation leads to this dense attention:

$$Attention(X) = W_p.attend(X, S) \quad (92)$$

where $W_p$ is the post attention matrix. Similarly, to implement factorized attention heads, one attention type is used alternatively per residual block or interleaved or a hyperparameter determines the ratio.

$$Attention(X) = W_p.attend(X, A^{(r \mod p)}) \quad (93)$$

where $r$ is the current residual block index and $p$ is the factorized headcount. An alternative merged head approach incorporates one head attend to target locations where both factorized heads would attend to. This approach is computationally more expensive by a constant factor.

$$Attention(X) = W_p.attend\left(X, \bigcup_{m=1}^{p} A^{(m)}\right) \quad (94)$$

A third alternative uses multiheaded attention, where attention products $(n_h)$ are parallelly computed and concatenated along the feature dimension.

$$Attention(X) = W_p(attend(X, A)^i)_{i \in \{1, --, n_h\}} \quad (95)$$

Multiple heads gave superior results whereas, for longer sequences where attention determines computation, sequential attention is preferred.

### IX-B REFORMER

Reformer reduces the Transformer attention complexity to $O(L \log L)$ via local sensitive hashing (LSH). This assigns each vector $x$ to a hash $h(x)$, where neighboring vectors obtain the same hash within hash buckets of similar size with high probability and remote ones do not. The modified LSH attention equation:

$$o_i = \sum_{j \in P_i} \exp\left(q_i.k_j - z(i, P_i)\right) v_j \quad (96)$$

where $P_i = \{j : i \geq j\}$

$P_i$ belongs to the set where $i^{th}$ position query attends to, $z$ is the partition function that contains a range of nearby keys to which a query attends to. For batching purposes, attention is performed over $\widetilde{P}_i = \{0, 1 ---l\} \supset P_i$ where $P_i$ is a subset of $\widetilde{P}_i$ and elements not in $P_i$ are masked.

$$o_i = \sum_{j \in \widetilde{P}_i} \exp(q_i.k_j - m(j, P_i) - z(i, P_i))v_j \quad (97)$$

$$\text{where } m(j, P_i) = \begin{cases} \infty, & \text{if } j \notin P_i \\ 0 & \text{otherwise} \end{cases}$$

Decoder implements masking to prevent access to future query positions. The set $P_i$ target items can only be attended by a query at $i^{th}$ position, by enabling attention within a single hash bucket. To further reduce the probability of similar items falling in different buckets, several parallel hashing $(n_{rounds})$ is performed with distinct hash functions $\{h^{(1)}, h^{(2)}, ..\}$

$$\mathcal{P}_i = \bigcup_{r=1}^{n_{rounds}} \mathcal{P}_i^r \text{ where } \mathcal{P}_i^r = \{j : h^{(r)}(q_i) = h^{(r)}(q_j)\} \quad (98)$$

Attention is done on chunks of sorted keys queries and keys to batch:

$$\widetilde{\mathcal{P}}_i^{(r)} = \left\{ j : \left\lceil \frac{s_i^{(r)}}{m} \right\rceil - 1 \leq \left\lceil \frac{s_j^{(r)}}{m} \right\rceil \leq \left\lceil \frac{s_i^{(r)}}{m} \right\rceil \right\} \quad (99)$$

From (96) and (97) we can write,

$$o_i = \sum_{j \in \widetilde{P}_i} \exp(q_i.k_j - m(j, P_i) - z(i, P_i))v_j \quad (100)$$

$$= \sum_{r=1}^{n_{rounds}} \exp(z(i, \mathcal{P}_i^{(r)} - z(i, \mathcal{P}_i)) \sum_{j \in \widetilde{P}_i^{(r)}} \frac{1}{N_{i,j}} \exp(q_i.k_j - m(j, \mathcal{P}_i^{(r)}) - z(i, \mathcal{P}_i^{(r)}))v_j \quad (101)$$

$$= \sum_{r=1}^{n_{rounds}} \exp(z(i, \mathcal{P}_i^{(r)} - z(i, \mathcal{P}_i)) o_i^{(r)} \quad (102)$$

$$o_i^{(r)} = \sum_{j \in \widetilde{P}_i^{(r)}} \exp(q_i.k_j - m(j, \mathcal{P}_i^{(r)}) - z(i, \mathcal{P}_i^{(r)}))v_j \quad (103)$$

The following example in figure 23 comprehensively demonstrates the various working mechanisms of the Reformer.

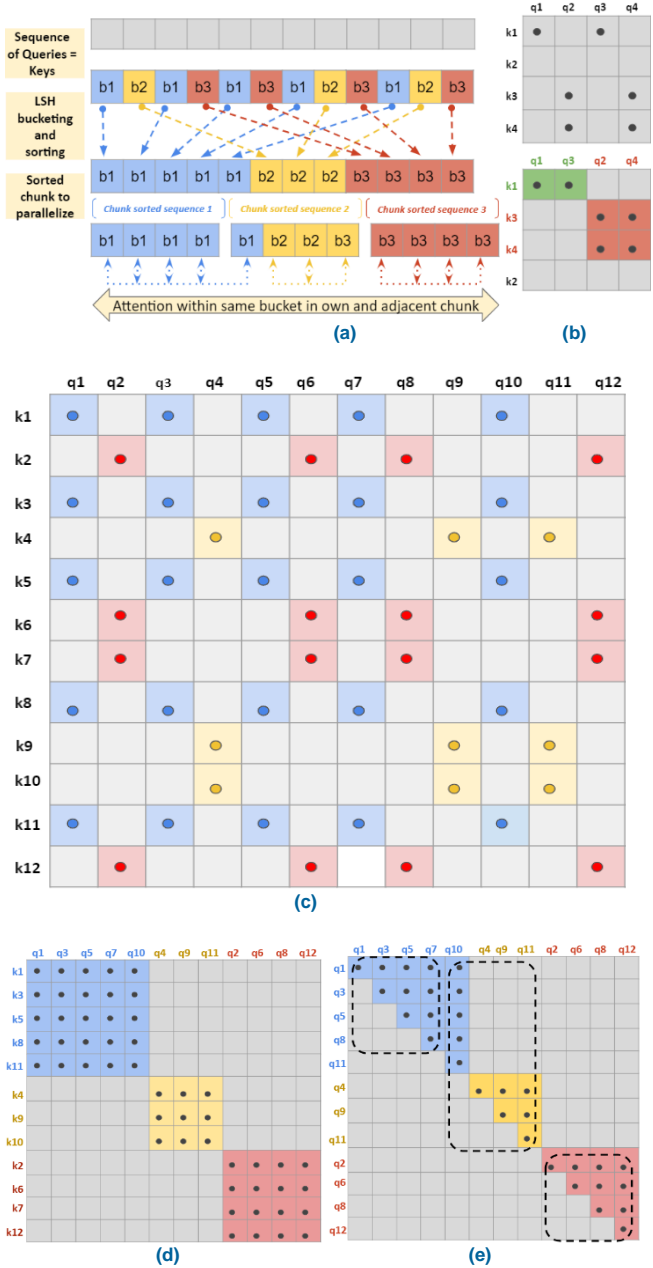

FIGURE 23. (a) Bucket formation of similar Attention vectors (b) Simple bucketing of a Query-Key pair (c) Query-Key sequence distribution based on (a) before Bucketing (d), (e) Bucketing and Chunking of (c)

Reversible Residual Networks [94] is another driving force behind Reformer's economical memory consumption where activation values are reconstructed on the fly during backpropagation excluding requirements to save activations in memory. From figure 24 below each layer's reversible block are recomputed from the next layer's activations as:

$$Y_1 = X_1 + f(X_{2\,Layer\,2}), Y_2 = X_2 + f(X_{1\,Layer\,1}) \quad (104)$$

$$X_1 = Y_1 - f(Y_{2\,Layer\,2}), X_2 = Y_2 - f(Y_{1\,Layer\,1}) \quad (105)$$

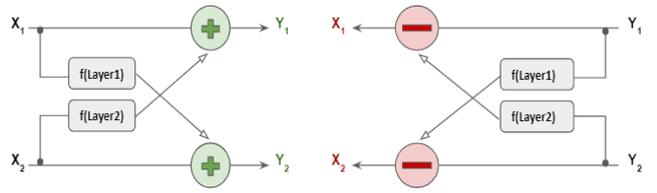

FIGURE 24. Rev-Nets skipping intermediate storage via precomputation

*IX-C A LITE BERT: ALBERT*

ALBERT, within a single model, integrates the following two-parameter reduction techniques that result in a mere 12M parameters as shown in figure 25. This results in almost 90% parameter reduction than BERT-base while maintaining competitive benchmark performances.

*(i)* <u>*Factorized Embedding Parameterization*</u>: For optimal results, NLP tasks require a large vocabulary $V$, where (embedding size) $E \equiv H$ (hidden layer) and embedding matrix $V \times E$ size can scale up to billion parameters. ALBERT factorizes the embedding space $E$ into two smaller matrices where embedding parameters are reduced from $O(V \times H)$ to $O(V \times E + E \times H)$.

*(ii)* <u>*Cross − Layer Parameter Sharing*</u>: ALBERT is built to share attention parameters across layers via a feed-forward network (FFN). Consequently, its inter-layer transitions were considerably smoother as results indicated weight sharing's stabilizing effect on network parameters.

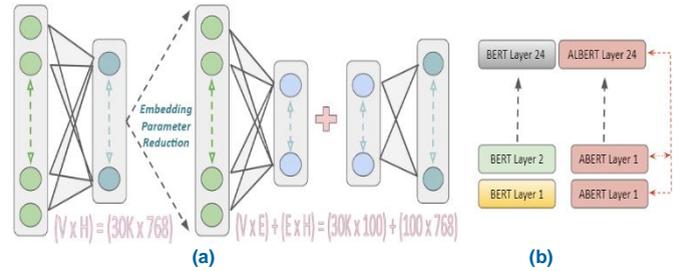

FIGURE 25. (a) Smaller model via Embedding size reduction (b) Effective learning via sharing of Attention parameters

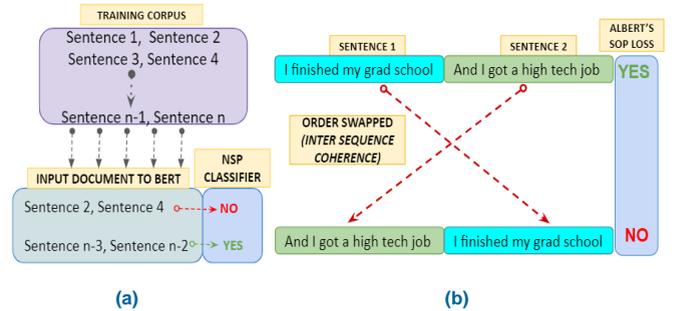

FIGURE 26. (a) BERT's NSP learning via simple non-reversed pair order (b) ALBERT's SOP dual sentiment learning via sentence order reversal.

Like BERT's NSP, ALBERT's sentence-order prediction (SOP) loss incorporated two-pronged learning from two positive successive text segments that also included its corresponding negative samples with orders reversed as demonstrated in figure 26. This influences the model to learn contextually the finer-grained discrepancies in any discourse giving superior coherent performances. Its MLM target implements $n$-gram masking that comprises up to 3-character sequences, like "World Cup Football" or "Natural Language Processing".

$$p(n) = \frac{1/n}{\sum_{k=1}^{n} 1/k} \tag{106}$$

### IX-D ELECTRA
The advantage lies in its contextual learning via effective discrimination, where it learns from all. input tokens unlike BERT's that learn from a mere 15% masked-out subset. ELCTRA implements "*replaced token detection*", as shown in figure 27, where contamination occurs by replacing few random tokens with probabilistic meaningful substitutions via Generator($G$), a small 'masked language model'.

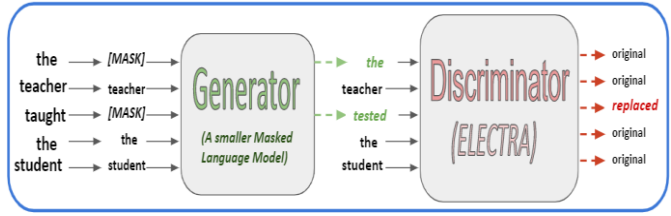

**FIGURE 27.** Replaced token detection via model's combined training

Simultaneously, via binary classification, a larger model Discriminator ($D$) is jointly pre-trained to predict if each token was restored correctly via the generator.

$$\mathcal{L}_{MLM}(x, \theta_G) = \mathbb{E}(\sum_{i \in m} -\log p_G(x_i | x^{masked})) \tag{107}$$

$$\mathcal{L}_{Disc}(x, \theta_D) = \tag{108}$$
$$\mathbb{E}\left(\sum_{t=1}^{n} -1(x_t^{corr} = x_t)\log D(x^{corr}, t) - 1(x_t^{corr} \neq x_t)\log(1 - D(x^{corr}, t))\right)$$

The two encoder-based networks ($G, D$) transform an input token sequence $x = [x_1, ..., x_n]$ into a contextualized vector representation $h_x = [h_1, ..., h_n]$. Via Softmax, $G$ yields the likelihood of generating a $t^{th}$ position token $x_t$, where $x_t = [MASK]$.

$$p_G(x_t | x) = \frac{\exp(e(x_t)^T h_G(x)_t)}{\sum_{x'} \exp(e(x')^T h_G(x)_t)} \tag{109}$$

The combined loss over a large corpus $\chi$ is minimized as:
$$\min_{\theta_G, \theta_D} \sum_{x \in \chi} \mathcal{L}_{MLM}(x, \theta_G) + \lambda \mathcal{L}_{Disc}(x, \theta_D) \tag{110}$$

### IX-E LINFORMER
It demonstrates [95] that attention weights are dominated by a few key entries, hence sequence length is down projected to a target output matrix via low-rank self-attention that achieves linear time and space complexity $O(1)$. During computation of keys and values, two linearly projected matrices are added $E_i, F_i \in \mathbb{R}^{n \times k}$, where $(n \times d)-$ dimensional key, value layers $KW_i^K$ and $VW_i^V$ are projected into $(k \times d)-$ dimensional key, value layers, thereafter resulting $(n \times k)-$ dimensional context mapping is computed using scaled dot-product attention.

$$head_i = \tag{111}$$
$$softmax\left(\frac{QW_i^Q(E_i KW_i^K)^T}{\sqrt{d_k}}\right)^{\overline{P}:n \times k} \cdot (F_i VW_i^V)^{k \times d}$$

If $k << n$, then a significant reduction of memory and space consumption is achieved. For further efficient optimization, parameter sharing between projections is performed at three levels: *(i) Headwise Sharing*: for each layer two projection matrices $E$ and $F$ are shared where $E_i = E, F_i = F$ through all heads $i$. *(ii) Key-Value Sharing*: including *(i)* key, value projections are shared where each layer's single projection matrix $E = E_i = F_i$ is created for each key-value projection matrix for all heads $i$ *(iii) Layer-wise Sharing*: a single projection matrix $E$ implemented for all layers, heads, keys, and values. For a 12-layer, 12-head Transformer, (i), (ii), (iii) will incorporate 24, 12, 1 distinct linear projection matrices, respectively.

### IX-E PERFORMER
The standard attention $(Q_{L \times d}. K_{d \times L}^T). V_{L \times d}$ results in quadratic time complexity of $O(L^2 d)$, preferable implementation of $Q_{L \times d}. (K_{d \times L}^T. V_{L \times d})$ leads to $O(d^2 L)$ where $L \gg d$. However, attention decomposition of query-key product into its pristine form is not possible after implementing the softmax non-linear function. However, pre softmax decomposition of attention is possible via approximation of lower-ranked queries and keys enabling greater efficiency, specifically $Q'K'^T \cong softmax(\frac{QK^T}{\sqrt{d}}) \cong \exp(QK^T)$. This is achieved via kernel approximation function $K(x, y) = \emptyset(x)^T \emptyset(y)$, the dot product of a high-dimensional feature map $\emptyset$. Contrary to the kernel trick where the dimensionality is increased, the Performer [96] decomposes the attention matrix $A(i, j) = K(q_i, k_j) = \exp(q_i, k_j^T)$ to a lower-dimensional feature map $\emptyset$.

## X. MODELING CLASSIFICATION OF LMs
Transformer based language models (LM) can be classified into 3 categories [97] from a modeling perspective:
- *(i) Autoregressive:* These are pre-trained feedforward models that predict future tokens from token history. Here output $y_t$ is dependent on the input at time instant $x_t$ and previous time step inputs $x_{<t}$. These are primarily decoder-based Transformers that incorporate causal masking where attention heads are prevented from attending to future tokens. Such models are generally fine-tuned for text generation purposes and deploy zero-shot learning in the GPT series.
- *(ii) Auto-Encoded:* These Encoder based models have full access to the input array, devoid of any masking. To learn they are pre-trained via incorporating input token masking schemes and then fine-tuned to reproduce the masked tokens as output. These models (BERT) are generally appropriate for sequence or token classification tasks.

(iii) *Sequence to Sequence*: These Encoder-Decoder-based generative models create data post learning from a massive dataset. Unlike discriminative distribution $P(Y|X)$, they model the joint distribution $P(X,Y)$ of input $X$ and target $Y$ where input can be corrupted on several schemes. Decoder-based causal masking is deployed to maximize learning for subsequent target generation. Models like BART and T5 perform best on NMT, summarization, or QA tasks.

A comprehensive overview of the above-mentioned modeling classification is presented in figure 29.

## XI. LANGUAGE MODEL PERFORMANCE COMPARISON

The quantitative performance of few major NLP models is shown in figure 28 that is based on the Glue and SuperGlue benchmarks. These benchmarks contain a variety of datasets that judge the model on several NLP tasks. With the highest number of trainable parameters, GPT-3 is the largest model in this comparison. Since GPT-3 is the newest model here, it does not participate in the older Glue benchmark.

From a qualitative perspective, the T5 within the same model uses the same loss function and hyperparameters spread across a variety of tasks leading to a multi-task learning environment. It performs the best as this scalable text to text generative (NLG) model couples the denoising objective during its training with massive amounts of unlabelled data. This leads to superior learning and greater generalized performances over NLU models like RoBERTa which are fine-tuned for individual downstream tasks after pre-training.

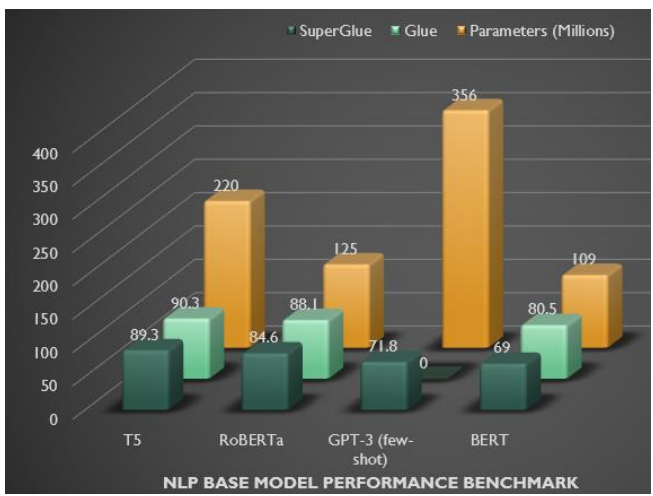

**FIGURE 28.** Graphical Representation of Language Model Performance

The primary motive of several rounds of fine-tuning in NLU models is to achieve strong performance on multiple tasks. The major disadvantages are the requirement for a new and typically large dataset for each task. This amplifies the potential for poor out-of-distribution generalization leading to unfair comparison with human-level abilities. GPT-3 does not operate on fine-tuning as its focus is to deliver task-agnostic execution. However, there is the scope of minimal fine-tuning in GPT-3 which leads to one or few-shot learning. The idea is to perform zero or minimal gradient updates post pre-training a huge model on a massive dataset. Though GPT-3 does not rank highly with the SuperGlue benchmark, the key is that this generative model is the quickest in learning any task at inference time. It matches performance with SOTA fine-tuned models on several NLP tasks in the zero, one, and few-shot settings. It also generates high-quality samples and gives a solid qualitative performance at tasks defined on the fly.

## XII. CONCLUSION AND FUTURE DIRECTIONS

We provide a comprehensive and detailed summary of the major language models that have led to the current SOTA in NLP performance. Since the launch of the Attention mechanism and Transformer architecture, NLP has advanced exponentially. We presented a high-level mind map of model classifications via a taxonomy. These classifications are primarily based on Transformer derivative architectures, built for specialized tasks like Language Understanding and Generation, Model Size Reduction via Distillation, Quantization and Pruning, Information Retrieval, Long Sequence Modeling, and other Generalized Model Reduction techniques. Recent language models are primarily driven by attaining higher NLP performance requiring huge computing resources. Thus, model scaling has been the natural pathway in industry. This exponential scaling coupled with higher attention complexity makes these models infeasible to access at a global scale. Subsequently, significant efforts have been made to engineer reasonably sized models and an efficient attention computation to speed up model convergence leading to lower latency in models.

Incorporating a Mixture of Expert (MoE) [98] methodology is an effective way for large models to achieve computational efficiency, as only a subset of the neural network is activated for every input. Consequently, this leads to sparsity, and although sparsity training is an active research area, current GPUs are better suited for dense matrix computations. While MoE models have demonstrated promise in training sparse matrices, their communication costs and complexity impede wide-scale deployment. Further, larger models are prone to memorize training data leading to overfitting and reduced learning [99]. To overcome this, models are only trained for a single epoch on de-duplicated instances on huge datasets, thereby exhibiting minimal overfitting.

Thus, MoE design coupled with a robust training paradigm in the future might lead to highly scalable and efficient models. These models will possess superior language understanding, as data memorization would be minimized. The current approach in SOTA models relies on supervised learning on huge datasets. A promising area of future enhancements in NLP would be incorporating reinforcement learning in Machine Translation, text summarization, and Q&A tasks.

| MODEL | DESCRIPTION | TASKS | LANGUGAE MODELING TYPE |
|---|---|---|---|
| GPT-I, II, III | • Unsupervised pre-training on large datasets<br>• Autoregressive Language Modeling and Causal Masking | Q&A, NMT, Reading Comprehension, Text Summarization, Common Sense Reasoning, Zero-Shot | *Autoregressive DECODER based Transformer* |
| XLNET | • Greater Contextual Learning via Factorized Ordering on Input's Sequence Length<br>• Bidirectional Contextual Language Modeling | Reading Comprehension, Natural Language Inference, Sentiment Analysis, Q&A | *Autoregressive DECODER based Transformer* |
| REFORMER | • Attention via Local Sensitive Hashing reducing memory footprint<br>• Incorporates re-computation of weights and activations bypassing their respective storage via reversible residual networks | Reduced Attention Complexity enabling lengthy sequence processing on pragmatic memory requirements | *Autoregressive DECODER based Transformer* |
| LONGFORMER | • Sparsity in Attention Matrices for lengthy sequence speedup and efficient computation<br>• Localized Attention for nearby tokens and all access Globalized Attention for few preselected tokens to enhance receptivity | Co-reference Resolution, Q&A, Document Classification. | *Autoregressive DECODER based Transformer* |
| BERT | • Deep Bidirectional Contextualization<br>• Masked Language Modeling (MLM) for continual learning | Sentence Classification, Q&A, Natural Language Inference, | *Auto-encoded ENCODER based Transformer* |
| RoBERTa | • Diverse learning via Dynamic Masking, where tokens are masked differently for each epoch<br>• Larger pre-training Batch-Size | Sentiment Analysis, Q&A, Natural Language Inference | *Auto-encoded ENCODER based Transformer* |
| DistilBERT | • Produces similar target probability distribution as its larger teacher model, BERT<br>• Generates cosine similarity between student and teacher model's hidden states | Semantic Textual Similarity, Semantic Relevance, Q&A, Textual Entailment | *Auto-encoded ENCODER based Transformer* |
| ALBERT | • Smaller and efficient model via Embedding Parameter Reduction, i.e., Factorized Parametrization<br>• Layers split into groups via Cross-Layer Parameter Sharing reducing memory footprint | Reading Comprehension, Semantic Textual Similarity, Q&A, Language Inference | *Auto-encoded ENCODER based Transformer* |
| ELECTRA | • Predict if the re-generated corrupted token is original or replaced via pre-training Generator<br>• Effective and low-cost discriminative learning via replaced token detection | Provides competitive performances on Sentiment Analysis, Natural Language Inference tasks at 25% compute | *Auto-encoded ENCODER based Transformer* |
| BART/mBART | • Superior sequence generation quality via greater noising variations<br>• Flexible denoising autoencoder acts as a language model in severest noising scheme | Supervised and Unsupervised multi-lingual Machine Translation, Q&A, Semantic Equivalence | *Generative Sequence to Sequence based Transformer* |
| T5/mT5 | • Positional Encodings learned at each layer for greater semantical performance<br>• Transforms all tasks in a text-to-text format to incorporate most NLP task varieties. | More Diverse and Challenging coreference, entailment, Q&A tasks via SuperGLUE benchmark | *Generative Sequence to Sequence based Transformer* |

**FIGURE 29.** Tabular Representation of Language Modeling Classification